\documentclass[letterpaper]{article} 
\usepackage[draft]{aaai25}  
\usepackage{times}  
\usepackage{helvet}  
\usepackage{courier}  
\usepackage[hyphens]{url}  
\usepackage{graphicx} 
\usepackage{natbib}  
\usepackage{caption} 
\usepackage{algorithm}
\usepackage{algorithmic}

\usepackage{newfloat}
\usepackage{listings}

\usepackage{amsmath,amsfonts}
\usepackage{algorithm}
\usepackage{algorithmic}
\usepackage{array}
\usepackage[caption=false,font=normalsize,labelfont=sf,textfont=sf]{subfig}
\usepackage{textcomp}
\usepackage{placeins}
\usepackage{url}
\usepackage{verbatim}
\usepackage{graphicx}
\usepackage{cite}
\usepackage{multicol}
\usepackage{multirow}
\usepackage{booktabs}
\usepackage{adjustbox}
\usepackage{enumitem}
\newcommand{\RomanNumeralCaps}[1]{\MakeUppercase{\romannumeral #1}}
\newcommand{\KL}{D_{\mathrm{KL}}}
\DeclareCaptionStyle{ruled}{labelfont=normalfont,labelsep=colon,strut=off} 
\floatstyle{ruled}
\newfloat{listing}{tb}{lst}{}
\floatname{listing}{Listing}
\title{Clustering Time Series Data with Gaussian Mixture Embeddings in a Graph Autoencoder Framework}
\author {
Amirabbas Afzali$^*$, Hesam Hosseini\footnote{
 Equal contribution. Author ordering is determined by coin flip.}, Mohmmadamin Mirzai , Arash Amini
}
\affiliations {
   \{amir8afzali, hesam8hosseini, moh.amin.mirzaii\}@gmail.com\\
   aamini@sharif.edu
}

\usepackage{bibentry}

\begin{document}

\maketitle

\begin{abstract}
Time series data analysis is prevalent across various domains, including finance, healthcare, and environmental monitoring. Traditional time series clustering methods often struggle to capture the complex temporal dependencies inherent in such data. In this paper, we propose the Variational Mixture Graph Autoencoder (VMGAE), a graph-based approach for time series clustering that leverages the structural advantages of graphs to capture enriched data relationships and produces Gaussian mixture embeddings for improved separability. Comparisons with baseline methods are included with experimental results, demonstrating that our method significantly outperforms state-of-the-art time-series clustering techniques. We further validate our method on real-world financial data, highlighting its practical applications in finance. By uncovering community structures in stock markets, our method provides deeper insights into stock relationships, benefiting market prediction, portfolio optimization, and risk management. 
\end{abstract}

%

\section{Introduction}
time series is commonly referred to a sequence of data points collected or recorded at successive time instances, usually at uniform intervals. For instance, in finance, time series data might include daily closing prices of a stock over a year\cite{shah2019stock}. In healthcare, it could be the EEG signal of a person's brain in a specific time interval \cite{siuly2016eeg}, and in environmental monitoring, it might involve hourly temperature readings \cite{zhao2009productivity}.

Numerous studies have been conducted on time series analysis, encompassing various tasks such as forecasting \cite{torres2021deep}, classification \cite{ismail2019deep}, clustering \cite{aghabozorgi2015time}, anomaly detection \cite{shaukat2021review}, visualization \cite{fang2020survey}, pattern recognition\cite{lin2012pattern}, and trend analysis \cite{mudelsee2019trend}.

Time series clustering is a powerful method for grouping similar time series data points based on their characteristics, especially when there is no prior knowledge of the data structure \cite{liao2005clustering}. It has diverse applications, such as stock market forecasting, where it is used for feature extraction to predict stock movements, helping investors anticipate market behavior and enhance model predictions \cite{babu2012clustering}. In portfolio optimization, clustering identifies stocks with similar traits, fostering diversification and reducing risks. Additionally, it supports risk management by predicting market volatility using algorithms like Kernel K-Means and Gaussian Mixture Models \cite{chaudhuri2016using} and contributes to fraud detection by flagging anomalies that deviate from typical cluster patterns \cite{close2020combining}.

Despite its practical significance, unsupervised time series clustering faces notable challenges. Time series data often vary significantly in their critical properties, features, temporal scales, and dimensionality across different domains. Real-world data further complicate this process by introducing issues such as temporal gaps and high-frequency noise \cite{hird2009noise}. To address these challenges, researchers have developed methods focusing on three main aspects: 1) time series similarity measures \cite{tan2020granger,li2021graph}, 2) discriminative representation learning \cite{ma2019learning,zhang2018salient,jorge2024time}, and 3) clustering mechanisms \cite{paparrizos2015k,li2022autoshape}. These advancements aim to enhance the reliability and applicability of time series clustering in complex, real-world scenarios.

In this paper, we leverage all these aspects by constructing a graph from time series data using dynamic time wrapping (DTW) \cite{Sakoe1978DynamicPA} to capture the relationships between individual time series. By exploiting a specialized graph autoencoder, we can also learn how to embed each node properly. This embedding not only represents the unique features of each data point but also captures shared features from nodes similar to the data point. To the best of our knowledge, this is the first work that employs graph autoencoder architecture for time series clustering.

The novel contributions of this work can be summarized as follows:

\begin{itemize}
\item We propose a new framework for time series clustering. This approach uses a graphical structure to capture more detailed information about data relationships. Turning a time series dataset into graphs can effectively capture both temporal and relational dependencies.
\item We introduce a specialized graph autoencoder, named Variational Mixture Graph Autoencoder (VMGAE), that generates a Mixture of Gaussian (MoG) embeddings. This allows the separation of data through a Gaussian Mixture Model (GMM) in the embedding space, enhancing clustering performance and providing a more precise representation of time series data.

\item We conducted a comprehensive comparison of our method against strong baselines, demonstrating significant improvements over the state-of-the-art. Additionally, we evaluated the practicality of our approach using real-world financial data.
\end{itemize}

\section{Related Work}
\noindent Time series data clustering has been a significant area of research for decades, leading to various algorithms. Classical clustering algorithms like k-means and spectral clustering can be executed on raw time series data, while some methods use modified versions of classical methods. K-shape \cite{paparrizos2015k}, assigns data to clusters based on their distance to centroids and updates the centroids like k-means, but instead uses cross-correlation for distance measurement. KSC(\cite{yang2011patterns}) uses k-means for clustering by adopting a
pairwise scaling distance measure and computing the spectral norm of a matrix for centroid computation.

Another approach is to use shapelets to extract discriminative features from time series data, as demonstrated in \cite{ulanova2015scalable}, \cite{zhang2018salient}, \cite{li2022autoshape}, and \cite{zhang2016unsupervised}. The main challenge in these methods is identifying suitable shapelets for the shapelet transform process, which extracts meaningful features from raw data to perform clustering. R-clustering method \cite{jorge2024time} employs random convolutional kernels for feature extraction, which are then used for clustering. Additionally, \cite{tan2020granger} implements a hierarchical clustering algorithm that uses Granger causality \cite{ding2006granger} as the distance measure, fusing pairs of data to create new time series and continuing the clustering process. STCN \cite{ma2020self} uses an RNN-based model to forecast time series data, employs pseudo labels for its classifier, and utilizes the learned features for clustering. Since our method leverages both the autoencoder architecture and a graph-based approach for clustering time series data, we will review autoencoder-based methods and graph-based techniques separately.

\subsection{Autoencoder-based methods} Autoencoders have demonstrated empirical success in clustering by using their learned latent features as data representations. For instance, DEC \cite{xie2016unsupervised} adds a KL-divergence term between two distributions to its loss function, alongside the reconstruction error loss, to make the latent space more suitable for clustering. DTC \cite{olive2020deep} introduces a new form of distribution for the KL-divergence term, applying this method to trajectory clustering. Another method, DCEC \cite{guo2017deep}, incorporates a convolutional neural network as its autoencoder within the DEC method for image clustering. VaDE \cite{jiang2016variational} adds a KL-divergence term between the Mixture-of-Gaussians prior and the posterior to the reconstruction loss. This is done using the embeddings of data points in the latent space of a variational autoencoder and a prior GMM distribution.
In the domain of time series clustering, DTCR \cite{ma2019learning} trains an autoencoder model with the addition of k-means loss on the latent space and employs fake data generation and a discriminator to classify real and fake data, enhancing the encoder's capabilities. Also, TMRC \cite{lee2024temporal} proposes a representation learning method called temporal multi-features representation learning (TMRL) to capture various temporal patterns embedded in time-series data and ensembles these features for time-series clustering. 
\subsection{Graph-based methods} Graphs have significantly enhanced the capabilities of deep learning methods in various tasks. Variational Graph Autoencoder (VGAE) \cite{kipf2016variational} utilizes GCN \cite{kipf2016semi} for link prediction and node classification tasks. Specifically, graphs have also found significant applications in the time series domain. Recent works such as \cite{song2020spatial}, \cite{cao2020spectral}, and \cite{yu2017spatio} use graph-based methods for time series forecasting, while \cite{zha2022towards} and \cite{xi2023lb} apply them for classification. Additionally, \cite{zhao2020multivariate}, \cite{deng2021graph}, and \cite{han2022learning} utilize graph-based models for anomaly detection in time series data. Graphs have also been employed for time series data clustering \cite{li2021graph}.

One of the critical challenges in graph-based methods for the time series domain is constructing the adjacency matrix.  Several methods address this issue by introducing metrics to compute the similarity or distance between two time series samples. The Granger causality method \cite{ding2006granger} leverages the causal effect of a pattern in one time series sample on another to measure similarity between samples. The Dynamic Time Warping (DTW) method \cite{Sakoe1978DynamicPA} minimizes the effects of shifting and distortion in time by allowing the elastic transformation of time series to compute the distance between two samples. There are many extensions of the DTW method, such as ACDTW \cite{li2020adaptively}, which uses a penalty function to reduce many-to-one and one-to-many matching, and shapeDTW \cite{zhao2018shapedtw}, which represents each temporal point by a shape descriptor that encodes structural information of local subsequences around that point and uses DTW to align two sequences of descriptors.

The similarity between two time series can be used directly as the edge representation, but the distance needs to be processed for use in the adjacency matrix. One method is to apply a threshold on distances to predict whether an edge exists between two nodes in a binary graph \cite{li2021graph}.

\section{Problem Definition and Framework}
\subsection{Notation}
\label{sec:notation}

In the following sections, we denote the training dataset as $\mathcal{D} = \{d_1, \dots, d_n\}$, where $d_i$ represents the $i$-th time series, and $n$ is the size of the training dataset. The length of the $i$-th time series is denoted by $l_i$. Each time series $d_i$ belongs to a cluster $c_i$, where $c_i \in \{1, \dots, K\}$, and $K$ is the number of clusters. Furthermore, we define $\mathbf{c} = [c_1, \dots, c_n]^{\top}$ as the vector of the corresponding clusters for all time series in $\mathcal{D}$.

Additionally, a graph is represented as $\mathbf{G} = \{\mathbf{V}, \mathbf{E}, \mathbf{X}\}$, where $\mathbf{V} = \{v_i\}_{i=1}^{n}$ is the set of nodes, and each edge $e_{i,j} = \langle v_i, v_j \rangle \in \mathbf{E}$ represents a connection between nodes $v_i$ and $v_j$. The structure of the graph is described by an adjacency matrix $\mathbf{A}$, where $\mathbf{A}_{i,j} = 1$ if $e_{i,j} \in \mathbf{E}$, and $\mathbf{A}_{i,j} = 0$ otherwise. The feature vector $\mathbf{x}_i \in \mathbf{X}$ corresponds to the content attributes of node $v_i$, which, in our context, is equivalent to the time series $d_i$.

Given the graph $\mathbf{G}$, our objective is to map each node $v_i \in \mathbf{V}$ to a low-dimensional vector $\mathbf{z}_i \in \mathbb{R}^h$. This mapping is formally defined as:

\footnotesize
\begin{equation*}
f : (\mathbf{A}, \mathbf{X}) \mapsto \mathbf{Z},
\end{equation*}
\normalsize

\noindent where the $i$-th row of the matrix $\mathbf{Z} \in \mathbb{R}^{n \times h}$ is denoted by $\mathbf{z}_i^{\top}$. Here, $h$ is the dimensionality of the embedding space. The matrix $\mathbf{Z}$, which contains these embeddings, is designed to preserve both the structural information of the graph, captured by $\mathbf{A}$ and the content features represented by $\mathbf{X}$.

\subsection{Overall Framework}
Our goal is to represent a time series dataset as a graph $\mathbf{G} = \{\mathbf{V}, \mathbf{E}, \mathbf{X}\}$, where each node corresponds to a time series. We aim to learn a robust embedding for each node in this graph to perform clustering. To achieve this, we first construct the graph. Next, we apply an unsupervised graph representation learning approach within an autoencoder framework, enhanced for clustering, to process the entire graph and learn effective node embeddings $\mathbf{Z}$.

\noindent \textbf{Graph Construction.}  Each time series is represented as a node in the graph construction phase. The distance matrix \( \mathbf{S} \), calculated using the Dynamic Time Warping (DTW) method, captures the alignment between time series of varying lengths. We then apply our novel transformation to convert these distances into similarity scores, which determine the graph's structure \( \mathbf{A} \). This approach ensures that the graph reflects the true underlying relationships in the data, preserving important temporal patterns.

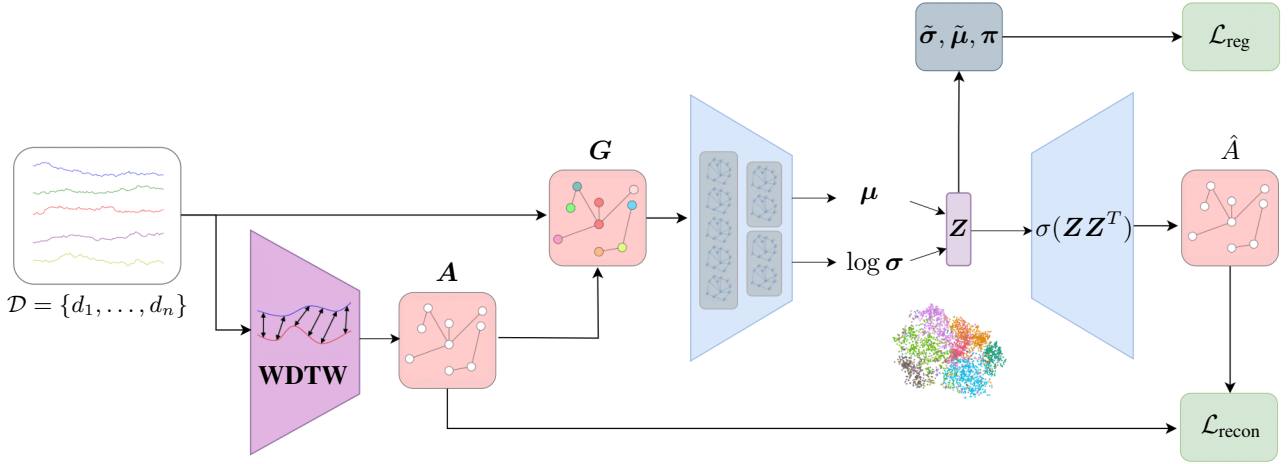
\begin{figure*}[!t]
    \centering
    \includegraphics[width=1\linewidth]{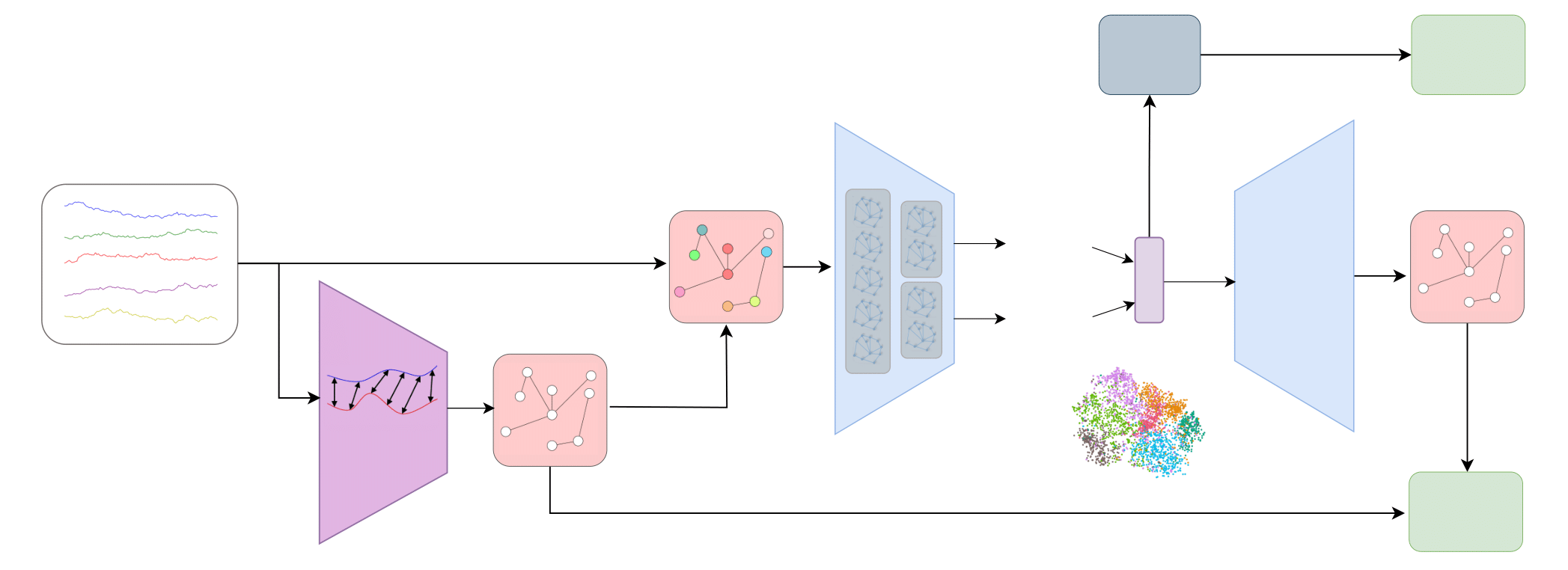}
    \put(-175,105){ $\boldsymbol{\mu}$}
    \put(-180,78){ $\log \boldsymbol{\sigma}$}
    \put(-333,74){\normalsize $\boldsymbol{A}$}
    \put(-36,120){\normalsize $\hat{A}$}
    \put(-275,120){\normalsize $\boldsymbol{G}$}
    \put(-400,35){\normalsize \textbf{WDTW}}
    \put(-495,62){\footnotesize $\mathcal{D} = \{d_1, \dots, d_n\}$}
    \put(-46,16.5){ $\mathcal{L}_{\text{recon}}$}
    \put(-43,164){ $\mathcal{L}_{\text{reg}}$}
    \put(-139,91){\footnotesize $\boldsymbol{Z}$}
    \put(-106,91){\normalsize $\sigma(\boldsymbol{ZZ}^T)$}
    \put(-150,164){\normalsize $\tilde{\boldsymbol{\sigma}}, \tilde{\boldsymbol{\mu}}, \boldsymbol{\pi}$}
    \caption{The general architecture of the Variational Mixture Graph Autoencoder (VMGAE). 
    The dataset \(\mathcal{D}\) consists of multiple time series data, and Weighted Dynamic Time Warping (WDTW) is used to compute distances that form the adjacency matrix \(\boldsymbol{A}\), representing connections in the graph \(\boldsymbol{G}\). 
    The mean \(\boldsymbol{\mu}\) and log standard deviation \(\log \boldsymbol{\sigma}\) are computed for the variational latent space, creating node embeddings \(\boldsymbol{Z}\).
    These embeddings undergo transformation to reconstruct the adjacency matrix \(\hat{\boldsymbol{A}}\), with the reconstruction loss \(\mathcal{L}_{\text{recon}}\) enforcing fidelity to \(\boldsymbol{A}\). The regularization loss \(\mathcal{L}_{\text{reg}}\) applies to the mixture model parameters \(\tilde{\boldsymbol{\sigma}}, \tilde{\boldsymbol{\mu}}, \boldsymbol{\pi}\), enhancing the latent space structure.}
    \label{fig:overal-architecture}
\end{figure*}

\noindent \textbf{Learning Representation via a Graph Structure.}  After constructing the graph, we leverage a specialized autoencoder framework to learn embeddings for each node. This unsupervised method compresses the graph’s information into a lower-dimensional representation \( \mathbf{Z} \). Our approach ensures that the resulting embeddings are informative, generalizable, and discriminative, making them particularly effective for clustering tasks. Figure \ref{fig:overal-architecture} provides a comprehensive overview of our method.

\section{Methodology}
This section describes the steps taken to represent a time series dataset as a graph and how we use this graph structure to learn meaningful embeddings for clustering. 

\subsection{Graph Construction}
For graph construction, we use a variant of DTW called Weighted Dynamic Time Warping (WDTW) and a constraint to limit the window size of the wrapping path.
The distance between two sequences \( X = (x_1, x_2, \ldots, x_N) \) and \( Y = (y_1, y_2, \ldots, y_M) \) with a weight funtion \( w \) and a window size \( W \) is computed as follows:

\footnotesize
\begin{equation}
    WDTW(X, Y) = \min_{\pi} \left( \sum_{(i,j) \in \pi} w[|i-j|] \cdot d_{\text{inner}}(x_i, y_j) \right),\label{eq:wdtw}
\end{equation}
\normalsize

\noindent subject to the constraint:

\footnotesize
\begin{equation}
|i - j| \leq W,
\end{equation}
\normalsize

\noindent where \( \pi \) is a warping path that aligns the sequences \( X \) and \( Y \), \( d_{\text{inner}}(x_i, y_j) \) is the distance between elements \( x_i \) and \( y_j \). This could be any customized distance. For simplicity, we use Euclidean distance. \( w[|i-j|] \) is a weight function that depends on the absolute difference between indices \( i \) and \( j \), and \( W \) is the window size that limits the maximum allowable shift between indices.

The weight function \( w[n] \) should be a monotonic function of \( n \), as it penalizes alignments where the indices are farther apart, favoring closer alignments. For simplicity, we set \( w[n] = \gamma \cdot n \), where \( \gamma \) is a positive hyperparameter.

\begin{figure}[ht!]
\centering
\includegraphics[width=2.6in]{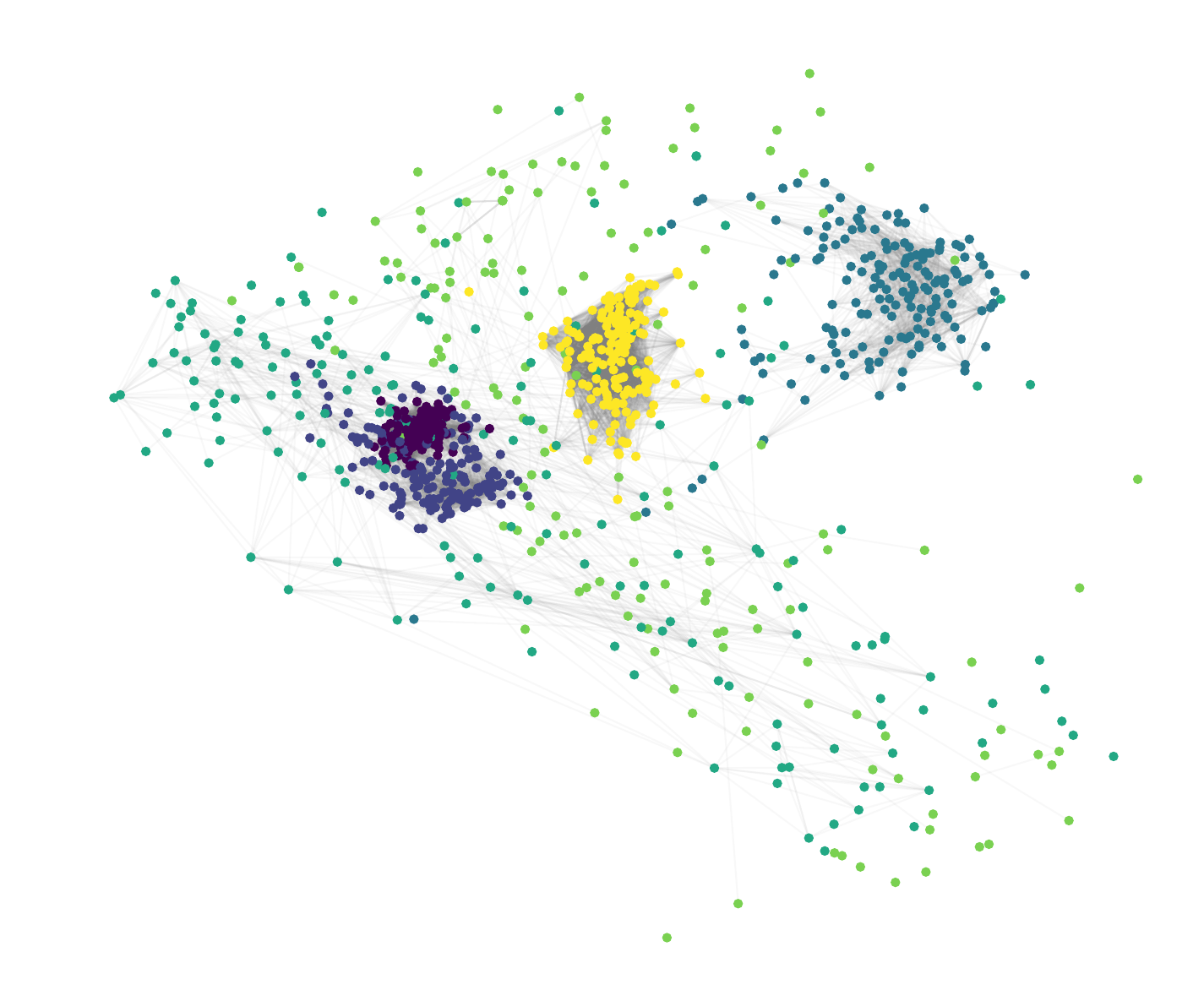}%
\caption{Graph visualizations of the Symbols dataset, illustrating effective data separation. Different colors correspond to distinct labels.}
\label{fig:graph_visualization}
\end{figure}

Given the training dataset \( \mathcal{D} = \{d_1, \dots, d_n\} \), we construct a distance matrix \( S \), where \( S_{ij} \) represents \( WDTW(d_i, d_j) \) with fixed parameters \( \gamma \) and \( W \). Next, we propose a novel transformation approach to convert the distance matrix \( S \) into a similarity matrix. By fixing the density rate \( \alpha = \frac{\#\{A_{ij} = 1\}}{n^2} \), we compute a threshold \( \delta \) to construct an adjacency matrix \( A \), where \( A_{ij} = 1 \) if \( S_{ij} < \delta \) and \( A_{ij} = 0 \) otherwise. The key difference compared to previous work \cite{li2021graph} is that, instead of fixing \( \delta \), we fix \( \alpha \) and use it to compute the corresponding \( \delta \) for each dataset. This is important because the optimal threshold \( \delta \) may vary across datasets, while the optimal \( \alpha \) is much more stable. Figure \ref{fig:graph_visualization} presents a sample graph constructed using this method. While the current representation demonstrates good separation, further refinement can be achieved with the use of \textit{VMGAE}.

\subsection{Learning Representation via a Graph Structure}
\label{sec:graph_learning}
\noindent \textbf{Graph Convolutional Autoencoder.}\quad In our unsupervised setting, we utilize a graph convolutional autoencoder architecture to embed a graph \( \mathbf{G} = \{ \mathbf{V}, \mathbf{E}, \mathbf{X} \} \) into a low-dimensional space. Specifically, we derive an embedding \( \mathbf{z_i} \in \mathbf{Z} \) for the \(i\)-th node of the graph. This approach presents two key challenges: 1) How can both the graph structure \( \mathbf{A} \) and node features \( \mathbf{X} \) be effectively integrated within the encoder? 2) What specific information should the decoder reconstruct?
\newline
\newline
\noindent\textbf{Graph Convolutional Layer.}\quad To effectively capture both the structural information \( \mathbf{A} \) and node features \( \mathbf{X} \) in a unified framework, we employ 
 graph convolutional network (GCN) \cite{kipf2016semi}. Graph convolutional operator extends the convolution operation to graph data in the spectral domain and applies a layer-wise transformation using a specialized convolutional function. Each layer of the graph convolutional network can be expressed as follows:

\footnotesize
\begin{equation}
\mathbf{Z}^{(l+1)} = \phi\left(\tilde{\mathbf{D}}^{-\frac{1}{2}} \tilde{\mathbf{A}} \tilde{\mathbf{D}}^{-\frac{1}{2}} \mathbf{Z}^{(l)} \mathbf{W}^{(l)} \right), 
\end{equation}
\normalsize

\noindent where \( \tilde{\mathbf{A}} = \mathbf{A} + \mathbf{I} \) and \( \tilde{\mathbf{D}}_{ii} = \sum_j \tilde{\mathbf{A}}_{ij} \). Here, \( \mathbf{I} \) represents the identity matrix and \( \phi \) is an activation function like ReLU \( \phi(x) = \max(0, x) \) . Also, \( \mathbf{Z}^{(l)} \) denotes the input at the \( l \)-th layer, and \( \mathbf{Z}^{(l+1)} \) is the output after the convolution operation. Initially, \( \mathbf{Z}^0 = \mathbf{X} \), where \( \mathbf{X} \in \mathbb{R}^{n \times m} \) represents the input features of the graph with \( n \) nodes and \( m \) features. The matrix \( \mathbf{W}^{(l)} \) contains the parameters to be learned. Additionally, in this work, we denote each convolutional layer with activation function $\phi$ as $f_{\phi}(\mathbf{Z}^{(l)},\mathbf{A} \mid \mathbf{W}^{(l)})$.
  
In addition to this convolutional layer, several other variants suitable for node-level tasks have been proposed \cite{veličković2018graphattentionnetworks, hamilton2018inductiverepresentationlearninglarge, du2018topologyadaptivegraphconvolutional, defferrard2017convolutionalneuralnetworksgraphs}. In Appendix F.2, we compare the effects of different convolutional layers on the performance of our method.
\newline
\newline
\noindent \textbf{Encoder Model} \( \mathcal{G}(\mathbf{X}, \mathbf{A}) \).\quad The encoder of \textit{VMGAE} is defined by an inference model:

\footnotesize
\begin{equation}
q(\mathbf{Z} | \mathbf{X}, \mathbf{A}) = \prod_{i=1}^{n} q(\mathbf{z}_i | \mathbf{X}, \mathbf{A}), 
\end{equation}
\normalsize

\footnotesize
\begin{equation}
q(\mathbf{z}_i | \mathbf{X}, \mathbf{A}) = \mathcal{N}(\mathbf{z}_i | \boldsymbol{\mu}_i, \text{diag}(\boldsymbol{\sigma}_i^2)).
\end{equation}
\normalsize

\noindent Here, $\boldsymbol{\mu}_i$ and $\log \boldsymbol{\sigma}_i$ are constructed using a two-layer convolutional network, where the weights $\mathbf{W}^{(0)}$ in the first layer are shared:

\footnotesize
\begin{equation} 
\mathbf{Z}^{(1)}=f_{\text{Relu}}(\mathbf{X}, \mathbf{A}|\mathbf{W}^{(0)}),
\end{equation}
\normalsize

\footnotesize
\begin{equation}
\begin{cases}
\boldsymbol{\mu} = f_{\text{linear}}(\mathbf{Z}^{(1)}, \mathbf{A}|\mathbf{W_{\mu}}^{(1)})
,\\ 
\log \boldsymbol{\sigma} = f_{\text{linear}}(\mathbf{Z}^{(1)}, \mathbf{A}|\mathbf{W_{\sigma}}^{(1)}).
\end{cases}
\end{equation}
\normalsize

\noindent The encoder model \( \mathcal{G}(\mathbf{X}, \mathbf{A}) \) encodes both graph structure and node features into a latent representation $\mathbf{Z} = q(\mathbf{Z} | \mathbf{X}, \mathbf{A})$. 
 According to the reparameterization trick, \(\mathbf{z}_i\) is obtained by:

 \footnotesize
 \begin{equation}
\mathbf{z}_i = \boldsymbol{\mu}_i + \boldsymbol{\sigma}_i \circ \boldsymbol{\epsilon}, 
\end{equation}
\normalsize
where \(\boldsymbol{\epsilon} \sim \mathcal{N}(0, I)\), \(\circ\) is element-wise multiplication.
\newline
\newline
\noindent \textbf{Decoder Model} \( \mathcal{D}(\mathbf{Z}, \mathbf{A}) \).\quad Decoder model is given by an inner product between latent variables:

\footnotesize
\begin{equation}
\label{eq:inner_dec}
p(\mathbf{A} \mid \mathbf{Z}) = \prod_{i=1}^{n} \prod_{j=1}^{n} p(A_{ij} \mid \mathbf{z}_i, \mathbf{z}_j) ,
\end{equation}
\normalsize

\noindent and the conditional probability is usually modeled as: 

\footnotesize
\begin{equation}
p(A_{ij} = 1 \mid \mathbf{z}_i, \mathbf{z}_j) = \sigma(\mathbf{z}_i^\top \mathbf{z}_j) ,
\end{equation}
\normalsize

\noindent where $\sigma(\cdot)$ is the logistic sigmoid function.

Thus, the embedding $\mathbf{Z}$ and the reconstructed graph $\hat{\mathbf{A}}$ can be presented as follows:

\footnotesize
\begin{equation}
\hat{\mathbf{A}} = \sigma(\mathbf{Z}\mathbf{Z}^\top), \quad \text{here } \mathbf{Z} = q(\mathbf{Z} \mid \mathbf{X}, \mathbf{A}) \quad \tag{9}
\end{equation}
\normalsize

\noindent \textbf{Learning Algorithm.}\quad In \textit{VMGAE}, our objective is to maximize the log-likelihood of the data points, \(\log p({\bf A})\). Based on the decoder model The joint probability $p(\mathbf{A}, \mathbf{Z}, \mathbf{c})$ can be factorized as:

\footnotesize
\begin{equation} \label{eq:fac}
    p(\mathbf{A}, \mathbf{Z}, \mathbf{c}) = p(\mathbf{A} \mid \mathbf{Z}) p(\mathbf{Z} \mid \mathbf{c}) p(\mathbf{c}).
\end{equation}
\normalsize

\noindent The log-likelihood can be expressed as:

\footnotesize
\begin{align}
    \log p({\bf A})&=\log \int_{\mathbf{Z}} \sum_{\textbf{c}} p({\bf A,Z},\textbf{c})\; d\mathbf{Z} \nonumber\\
    &\geq \mathbb{E}_{q({\bf Z}, \mathbf{c} \mid {\bf X}, {\bf A})} \left[ \log \frac{p({\bf A}, {\bf Z}, \mathbf{c})}{q({\bf Z}, \mathbf{c} \mid {\bf X}, {\bf A})} \right] \label{eq:jenson} \\
    &= \mathcal{L}_{\textup{ELBO}}({\bf X}, {\bf A}). \label{eqn:loglikelihood}
\end{align}
\normalsize

\noindent The inequality in Equation \ref{eq:jenson} is derived from Jensen's inequality. Instead of maximizing the log-likelihood directly, we aim to maximize its Evidence Lower Bound (ELBO), and using the factorization in Equation \ref{eq:fac}, it can be rewritten as follows:

\footnotesize
\begin{align}
&\mathcal{L}_{\textup{ELBO}}({\bf X,A})=E_{q({\bf Z},\mathbf{c}|{\bf X,A})}[\log p({\bf A,Z},\mathbf{c}) -\log q({\bf Z},\mathbf{c}|{\bf X,A})]\nonumber \\
& =E_{q({\bf Z},\mathbf{c}|{\bf X,A})}\left[\log p({\bf A|Z}) + \log p({\bf Z}|\mathbf{c})+\log p(\mathbf{c})\right]\nonumber\\
&\hspace{4mm}-E_{q({\bf Z},\mathbf{c}|{\bf X,A})}\left[\log q({\bf Z|X,A}) + \log q(\mathbf{c}|{\bf X,A})\right],
\label{eq:ELBO}
\end{align}
\normalsize

\noindent where the last line is obtained under the assumption of a mean-field distribution for \(q({\bf Z},\mathbf{c}|{\bf X,A})\).

\quad Similar to the approach in \cite{jiang2016variational}, a mixture of Gaussian latent variables is used to learn the following distributions:

\footnotesize
\begin{align}
p(c_i) & = \operatorname{Cat}(c_i \mid \boldsymbol{\pi}) \\
p(\mathbf{z_i} \mid c_i). & =\mathcal{N}\left(\mathbf{z_i} \mid \tilde{\boldsymbol{\mu}}_{c_i}, \tilde{\boldsymbol{\sigma}}_{c_i}^2 \mathbf{I}\right). 
\end{align}
\normalsize

\noindent By assuming a mean-field distribution, the joint probability \(p(\mathbf{c})\) and \(p(\mathbf{Z} \mid \mathbf{c})\) can be factorized as:

\footnotesize
\begin{align}
p(\mathbf{c}) & =\prod_{i=1}^n\operatorname{Cat}(c_i \mid \boldsymbol{\pi}), \label{eq:joint_cat}\\
p(\mathbf{Z} \mid \mathbf{c}) & = \prod_{i=1}^n\mathcal{N}\left(\mathbf{z_i} \mid \tilde{\boldsymbol{\mu}}_{c_i}, \tilde{\boldsymbol{\sigma}}_{c_i}^2 \mathbf{I}\right), \label{eq:joint_normal}
\end{align}
\normalsize

\noindent where $\boldsymbol{\pi}_k$ is the prior distribution of cluster $k$ hence $\sum_k \boldsymbol{\pi}_k=1$,
$\operatorname{Cat}(\cdot \mid \boldsymbol{\pi})$ is the categorical distribution parametrized by \(\boldsymbol{\pi}\). Also \(\boldsymbol{\mu}_{c_i}\)
 Moreover, $\boldsymbol{\sigma}_{c_i}^2$ are the mean and the variance of the Gaussian distribution corresponding to cluster $c_i$, $\mathbf{I}$ is an identity matrix.

Using Monte Carlo estimation for the expected value calculation in full-batch mode and substituting the assumptions from Equations \ref{eq:joint_cat} and \ref{eq:joint_normal} into Equation \ref{eq:ELBO}, the objective function can be expressed as:

\footnotesize
\begin{align}
\label{eq:elbow}
&\hspace{2mm} \mathcal{L}_{\text {ELBO }} (\mathbf{X},\mathbf{A})= \nonumber \\ 
&\hspace{2mm}\frac{1}{n}\sum_{i=1}^n\bigg[\frac{1}{n}\bigg(\sum_{j=1}^n A_{ij}\log{\hat{A}_{ij}}+(1-A_{ij})\log(1-\hat{A}_{ij})\bigg)\nonumber \\
& \hspace{-1mm} -\frac{1}{2} \sum_{c_i=1}^K q(c_i \mid \mathbf{X},\mathbf{A}) \sum_{h=1}^H\bigg( \log \tilde{\boldsymbol{\sigma}}_{c_i h}^2+\frac{{\boldsymbol{\sigma}}_h^2}{\tilde{\boldsymbol{\sigma}}_{c_i h}^2} + \frac{\left({\boldsymbol{\mu}}_h-\tilde{\boldsymbol{\mu}}_{c_i h}\right)^2}{\tilde{\boldsymbol{\sigma}}_{c_i h}^2}\bigg) \nonumber\\
&\hspace{-1mm} + \sum_{c_i=1}^K q(c_i \mid \mathbf{X},\mathbf{A}) \log \frac{\boldsymbol{\pi}_{c_i}}{q(c_i \mid \mathbf{X},\mathbf{A})}+\frac{1}{2} \sum_{h=1}^H\left(1+\log {\boldsymbol{\boldsymbol{\sigma}}_h^2}\right) \bigg].
\end{align}
\normalsize

The first term represents the standard reconstruction loss, while the second and third terms act as regularizers, encouraging the model to generate a Gaussian mixture embedding. A detailed derivation of Equation \ref{eq:elbow} is provided in Appendix A.

The next question is how to compute $q(c_i\mid \mathbf{X},\mathbf{A})$. According to the derivations in Appendix B, the ELBO can be rewritten as:

\footnotesize
\begin{align}
 &\mathcal{L}_{\mathrm{ELBO}}(\mathbf{X},\mathbf{A})=E_{q(\mathbf{Z}, \mathbf{c} \mid \mathbf{X},\mathbf{A})}\left[\log \frac{p(\mathbf{A}, \mathbf{Z}, \mathbf{c})}{q(\mathbf{Z}, \mathbf{c} \mid \mathbf{X},\mathbf{A})}\right] \nonumber\\
 &\hspace{4mm} = E_{q(\mathbf{Z} \mid \mathbf{X},\mathbf{A})}\left[\log \frac{p(\mathbf{A}\mid \mathbf{Z})p(\mathbf{Z})}{q(\mathbf{Z}\mid \mathbf{X},\mathbf{A})}\right]\nonumber\\
&\hspace{-4mm} -\sum_{i=1}^n\int_{\mathbf{z_i}} q(\mathbf{z_i} \mid \mathbf{X},\mathbf{A}) \KL (q(c_i \mid \mathbf{X},\mathbf{A}) \| p(c_i \mid \mathbf{z_i})) d \mathbf{z_i}.
\end{align}
\normalsize

\noindent In the equation above, the first term is independent of $\mathbf{c}$, and the second term is non-negative. Therefore, similar to the approach in \cite{jiang2016variational}, to maximize $\mathcal{L}_{\text{ELBO}}(\mathbf{X},\mathbf{A})$, we assume $\KL(q(c_i | \mathbf{X},\mathbf{A}) \| p(c_i | \mathbf{z_i}))$ to be zero. Consequently, the following equation can be used to compute $q(c_i | \mathbf{X},\mathbf{A})$ in \textit{VMGAE}:

\footnotesize
\begin{equation}
q(c_i \mid \mathbf{X},\mathbf{A})=p(c_i \mid \mathbf{z_i}) \equiv \frac{p(c_i) p(\mathbf{z_i} \mid c_i)}{\sum_{c^{\prime}=1}^K p\left(c^{\prime}\right) p\left(\mathbf{z_i} \mid c^{\prime}\right)}.
\end{equation}
\normalsize

\noindent While the learned distribution can be directly used for clustering, we have empirically found that refitting a GMM on the learned representation \( q(\mathbf{Z} | \mathbf{X},\mathbf{A}) \) significantly improves clustering performance.

\begin{table*}[htbp]
    \centering
    \adjustbox{max width=0.8\textwidth}{
    \begin{tabular}{lccccccccccccccccccccc}
        \toprule
        Dataset & K-means & SC & KSC & K-shape & u-shaplet & USSL & DTCR & STCN & R-clust &\multicolumn{1}{c|}{TMRC} & VMGAE \\
        \midrule
        Beef & 0.2925 & 0.4063 & 0.3828 & 0.3338 & 0.3413 & 0.3338 & 0.5473 & 0.5432  & 0.2475 &\multicolumn{1}{c|}{\textbf{0.7424}} & 0.5237\\
        Car & 0.2540 & 0.3349 & 0.2719 & 0.3771 & 0.3655 & 0.4650 & 0.5021 & 0.5701 & 0.5390 & \multicolumn{1}{c|}{0.3917} & \textbf{0.6193} \\
        DiatomSizeReduction & 0.9300 & 0.8387 & \textbf{1.0000} & \textbf{1.0000} & 0.4849 & \textbf{1.0000} & 0.9418 & \textbf{1.0000} & 0.6154 & \multicolumn{1}{c|}{0.6324} & 0.8882 \\
        Dist.Phal.Outl.AgeGroup & 0.1880 & 0.3474 & 0.3331 & 0.2911 & 0.2577 & 0.3846 & 0.4553 & \textbf{0.5037} & 0.4343 & \multicolumn{1}{c|}{0.3298} & 0.4400 \\
        ECG200 & 0.1403 & 0.1350 & 0.1403 & 0.3682 & 0.1323 & 0.3776 & 0.3691 & \textbf{0.4316} & 0.1561 & \multicolumn{1}{c|}{0.3763} & 0.3643 \\
        ECGFiveDays & 0.0002 & 0.0005 & 0.0682 & 0.0002 & 0.1498 & 0.6502 & 0.8056 & 0.3582 & 0.0173 & \multicolumn{1}{c|}{0.2758} & \textbf{0.8378}\\
        Meat & 0.2510 & 0.2732 & 0.2846 & 0.2254 & 0.2716 & 0.9085 & 0.9653 & 0.9393 & 0.6420 & \multicolumn{1}{c|}{0.7980} & \textbf{1.0000}\\
        Mid.Phal.TW & 0.4134 & 0.4952 & 0.4486 & 0.5229 & 0.4065 & \textbf{0.9202} & 0.5503 & 0.6169 & 0.4138 & \multicolumn{1}{c|}{0.4802} & 0.4409\\
        OSULeaf & 0.0208 & 0.0814 & 0.0421 & 0.0126 & 0.0203 & 0.3353 & 0.2599 & 0.3544 & \textbf{0.4453} & \multicolumn{1}{c|}{0.3012} & 0.3739\\
        Plane & 0.8598 & 0.9295 & 0.9218 & 0.9642 & \textbf{1.0000} & \textbf{1.0000} & 0.9296 & 0.9615 & 0.9892 & \multicolumn{1}{c|}{0.8917} & 0.9678\\
        Prox.Phal.Outl.AgeGroup & 0.0635 & 0.4222 & 0.0682 & 0.0110 & 0.0332 & \textbf{0.6813} & 0.5581 & 0.6317 & 0.5665 & \multicolumn{1}{c|}{0.5731} & 0.5639\\
        SonyAIBORobotSurface & 0.6112 & 0.2564 & 0.6129 & 0.7107 & 0.5803 & 0.5597 & 0.6634 & 0.6112 & 0.6620 & \multicolumn{1}{c|}{0.2300} & \textbf{0.9319}\\
        SwedishLeaf & 0.0168 & 0.0698 & 0.0073 & 0.1041 & 0.3456 & \textbf{0.9186} & 0.6663 & 0.6106 & 0.7151 & \multicolumn{1}{c|}{0.5099} & 0.5886\\
        Symbols & 0.7780 & 0.7855 & 0.8264 & 0.6366 & 0.8691 & 0.8821 & 0.8989 & 0.8940 & 0.8775 & \multicolumn{1}{c|}{0.8159} & \textbf{0.8996}\\
        ToeSegmentation1 & 0.0022 & 0.0353 & 0.0202 & 0.3073 & 0.3073 & 0.3351 & 0.3115 & 0.3671 & 0.0179 & \multicolumn{1}{c|}{\textbf{1.0000}} & 0.3081\\
        TwoPatterns & 0.4696 & 0.4622 & 0.4705 & 0.3949 & 0.2979 & 0.4911 & 0.4713 & 0.4110 & 0.3181 & \multicolumn{1}{c|}{0.1347} & \textbf{1.0000}\\
        TwoLeadECG & 0.0000 & 0.0031 & 0.0011 & 0.0000 & 0.0529 & 0.5471 & 0.4614 & 0.6911 & 0.4966 & \multicolumn{1}{c|}{0.0287} & \textbf{0.8726}\\
        Wafer & 0.0010 & 0.0010 & 0.0010 & 0.0010 & 0.0010 & 0.0492 & 0.0228 & 0.2089 & 0.0000 & \multicolumn{1}{c|}{\textbf{0.5019}} & 0.2136\\
        WordSynonyms & 0.5435 & 0.4236 & 0.4874 & 0.4154 & 0.3933 & 0.4984 & 0.5448 & 0.3947 & \textbf{0.8885} & \multicolumn{1}{c|}{0.4210} & 0.5812\\
        \midrule
        \textbf{AVG Rank} & 8.7894 & 7.6316 & 7.1053 & 7.4739 & 8.1053 & 3.4736 & 3.7368 & 3.4210 & 5.8947 & \multicolumn{1}{c|}{6.0000} & \textbf{3.1579} \\
        \textbf{AVG NMI} & 0.3071 & 0.3316 & 0.3362 & 0.3513 & 0.3321 & 0.5967 & 0.5749 & 0.5841 & 0.4759 & \multicolumn{1}{c|}{0.4965} & \textbf{0.6553} \\
        \textbf{Best} & 0 & 0 & 1 & 1 & 1 & 5 & 0 & 3 & 2 & \multicolumn{1}{c|}{3} & \textbf{7}\\
        \bottomrule
    \end{tabular}
    }
    \caption{Normalized Mutual Information (NMI) comparisons on 19 time series datasets }
    \label{tab:nmi_comparison}
\end{table*}

Additionally, in our experiments, we introduce a weight parameter $\lambda$ for the second component of the loss function, allowing us to balance the contribution of each term in the final loss function:

\footnotesize
\begin{align}
\label{eq:final_loss}
\mathcal{L}_{\text{VMGAE}}(\mathbf{X}, \mathbf{A}) = \mathcal{L}_{\text{recon}}(\mathbf{A}) + \lambda \cdot \mathcal{L}_{\text{reg}}(\mathbf{X}, \mathbf{A}),
\end{align}
\normalsize

\noindent where \(\mathcal{L}_{\text{recon}}(\mathbf{A})\) represents the reconstruction loss on the adjacency matrix \(\mathbf{A}\):

\footnotesize
\begin{align}
\label{eq:recon_loss}
\mathcal{L}_{\text{recon}}(\mathbf{A}) = -\frac{1}{n^2} \sum_{i,j=1}^n \bigg( A_{ij} \log{\hat{A}_{ij}} + (1 - A_{ij}) \log(1 - \hat{A}_{ij}) \bigg),
\end{align}
\normalsize

\noindent and the regularizer term \(\mathcal{L}_{\text{reg}}(\mathbf{X}, \mathbf{A})\) is defined as follows:

\footnotesize
\begin{align}
&\mathcal{L}_{\text {reg}}(\mathbf{X},\mathbf{A})=\frac{1}{n} \sum_{i=1}^n \bigg[\frac{1}{2} \sum_{c_i=1}^K q(c_i \mid \mathbf{X},\mathbf{A}) \sum_{h=1}^H\bigg( \log \tilde{\boldsymbol{\sigma}}_{c_i h}^2+\frac{{\boldsymbol{\sigma}}_h^2}{\tilde{\boldsymbol{\sigma}}_{c_i h}^2} \nonumber\\
&\hspace{4mm}+\frac{\left({\boldsymbol{\mu}}_h-\tilde{\boldsymbol{\mu}}_{c_i h}\right)^2}{\tilde{\boldsymbol{\sigma}}_{c_i h}^2}\bigg) - \sum_{c_i=1}^K q(c_i \mid \mathbf{X},\mathbf{A}) \log \frac{\boldsymbol{\pi}_{c_i}}{q(c_i \mid \mathbf{X},\mathbf{A})} \nonumber\\
&\hspace{4mm} +\frac{1}{2} \sum_{h=1}^H\left(1+\log {\boldsymbol{\sigma}_h^2}\right)\bigg) \bigg].
\end{align}
\normalsize

Compared to vanilla GAE and VGAE, our method introduces only a few additional parameters 
\(\tilde{\boldsymbol{\mu}}, \tilde{\boldsymbol{\sigma}}, \text{ and } \boldsymbol{\pi}\), which need to be learned. 
However, this does not significantly increase the computational overhead. Initializing these parameters using a GMM proves effective. In practice, performing a few epochs of pretraining with GAE—e.g., using only the reconstruction loss—followed by fitting a GMM on the latent embeddings is sufficient to achieve a strong initialization. 

Finally, we summarize the complete set of steps involved in our proposed method in Algorithm \ref{alg:VMGAE}.

        
        
        
        
        
        

\begin{table*}[htbp]
    \centering
    \adjustbox{max width=0.8\textwidth}{
    \begin{tabular}{lccccccccccccccccccccc}
        \toprule
        Dataset & K-means & SC & KSC & K-shape & u-shaplet & USSL & DTCR & STCN & R-Clust &\multicolumn{1}{c|}{TMRC} & VMGAE \\
        \midrule
        Beef & 0.6713 & 0.6206 & 0.7057 & 0.5402 & 0.6966 & 0.6966 & 0.8046 & 0.7471 & 0.6703 & \multicolumn{1}{c|}{\textbf{0.8229}} & 0.7862\\
        Car & 0.6345 & 0.6621 & 0.6898 & 0.7028 & 0.6418 & 0.7345 & 0.7501 & 0.7372 & 0.7507 & \multicolumn{1}{c|}{0.7322} & \textbf{0.8045} \\
        DiatomSizeReduction & 0.9583 & 0.9254 & \textbf{1.0000} & \textbf{1.0000} & 0.7083 & \textbf{1.0000} & 0.9682 & 0.9921 & 0.8140 & \multicolumn{1}{c|}{0.8539} & 0.9719 \\
        Dist.Phal.Outl.AgeGroup & 0.6171 & 0.7278 & 0.6535 & 0.6020 & 0.6273 & 0.6650 & \textbf{0.7825} & \textbf{0.7825} & 0.7425 & \multicolumn{1}{c|}{0.6477} & 0.6827\\
        ECG200 & 0.6315 & 0.5078 & 0.6315 & 0.7018 & 0.5758 & 0.7285 & 0.6648 & 0.7018 & 0.6206 & \multicolumn{1}{c|}{0.7424} & \textbf{0.7862} \\
        ECGFiveDays & 0.4783 & 0.4994 & 0.5257 & 0.5020 & 0.5968 & 0.8340 & \textbf{0.9638} & 0.6504 & 0.0173 & \multicolumn{1}{c|}{0.6492} & 0.9523\\
        Meat & 0.6595 & 0.7197 & 0.6723 & 0.6575 & 0.6742 & 0.7740 & 0.9763 & 0.9186 & 0.8341 & \multicolumn{1}{c|}{0.8847} & \textbf{1.0000}\\
        Mid.Phal.TW & 0.0983 & 0.8052 & 0.8187 & 0.6213 & 0.7920 & 0.7920 & \textbf{0.8638} & 0.8625 & 0.7915 & \multicolumn{1}{c|}{0.6850} & 0.8132\\
        OSULeaf & 0.5615 & 0.7314 & 0.5714 & 0.5538 & 0.5525 & 0.6551 & 0.7739 & 0.7615 & \textbf{0.8067} & \multicolumn{1}{c|}{0.7644} & 0.7798\\
        Plane & 0.9081 & 0.9333 & 0.9603 & 0.9901 & \textbf{1.0000} & \textbf{1.0000} & 0.9549 & 0.9663 & 0.9973 & \multicolumn{1}{c|}{0.9472} & 0.9868\\
        Prox.Phal.Outl.AgeGroup & 0.5288 & 0.7791 & 0.5305 & 0.5617 & 0.5206 & 0.7939 & 0.8091 & \textbf{0.8379} & 0.8021 & \multicolumn{1}{c|}{0.8189} &  0.8147\\
        SonyAIBORobotSurface & 0.7721 & 0.5082 & 0.7726 & 0.8084 & 0.7639 & 0.8105 & 0.8769 & 0.7356 & 0.8843 & \multicolumn{1}{c|}{0.6529} & \textbf{0.9834}\\
        SwedishLeaf & 0.4987 & 0.6897 & 0.4923 & 0.5333 & 0.6154 & 0.8547 & 0.9223 & 0.8872 & \textbf{0.9302} & \multicolumn{1}{c|}{0.8537} & 0.8825\\
        Symbols & 0.8810 & 0.8959 & 0.8982 & 0.8373 & 0.9603 & 0.9200 & 0.9168 & 0.9088 & \textbf{0.9821} &\multicolumn{1}{c|}{0.9088} & 0.9677\\
        ToeSegmentation1 & 0.4873 & 0.4996 & 0.5000 & 0.6143 & 0.5873 & 0.6718 & 0.5659 & 0.8177 & 0.5112 & \multicolumn{1}{c|}{\textbf{1.0000}} & 0.6712 \\
        TwoPatterns & 0.8529 & 0.6297 & 0.8585 & 0.8046 & 0.7757 & 0.8318 & 0.6984 & 0.7619 & 0.7273 & \multicolumn{1}{c|}{0.6295} & \textbf{1.0000}\\
        TwoLeadECG & 0.5476 & 0.5018 & 0.5464 & 0.8246 & 0.5404 & 0.8628 & 0.7114 & 0.9486 & 0.7984 & \multicolumn{1}{c|}{0.5873} & \textbf{0.9655}\\
        Wafer & 0.4925 & 0.5336 & 0.4925 & 0.4925 & 0.4925 & 0.8246 & 0.7338 & 0.8433 & 0.5349 & \multicolumn{1}{c|}{\textbf{0.9082}} & 0.5853\\
        WordSynonyms & 0.8775 & 0.8647 & 0.8727 & 0.7844 & 0.8230 & 0.8540 & 0.8984 & 0.8748 & 0.8995 & \multicolumn{1}{c|}{0.8875} & \textbf{0.9168}\\
        \midrule
        \textbf{AVG Rank} & 8.7368 & 8.3684 & 6.8947 & 7.5263 & 7.7895 & 4.4739 & 4.0000 & 3.9474 & 5.2105 & \multicolumn{1}{c|}{5.5263} & \textbf{2.6842} \\
        \textbf{AVG RI} & 0.6398 & 0.6860 & 0.6943 & 0.6911 & 0.6812 & 0.8054 & 0.8229 & 0.8282 & 0.7428  & \multicolumn{1}{c|}{0.7882} & \textbf{0.8605} \\
        \textbf{Best} & 0 & 0 & 1 & 1 & 1 & 2 & 3 & 2 & 3 &\multicolumn{1}{c|}{3} & \textbf{7}\\
        \bottomrule
    \end{tabular}
    }
        \caption{Rand Index (RI) comparisons on 19 time series datasets }
    \label{tab:ri_comparison}
\end{table*}

\begin{algorithm}[tb]
\caption{VMGAE Training Procedure}
\label{alg:VMGAE}
\footnotesize
\textbf{Input}: Time series dataset $\mathcal{D}$ \\
\textbf{Parameters}: Hyperparameters \(\{ W, \gamma, \lambda, \alpha \}\), Pre-training iterations $T_{\text{pre}}$, Training iterations $T$ \\
\textbf{Output}: Clustering results
\begin{algorithmic}[1] 
\STATE Compute the distance matrix $S$ using WDTW (Eq.~\eqref{eq:wdtw}).
\STATE Convert the distance matrix $S$ into an adjacency matrix $\mathbf{A}$.
\STATE Initialize GAE with random weights.
\FOR{$t = 1$ to $T_{\text{pre}}$}
    \STATE Pre-train the GAE by minimizing $\mathcal{L}_{\text{recon}}(\mathbf{A})$ (Eq.~\eqref{eq:recon_loss}).
\ENDFOR
\STATE Fit a GMM to the latent representations $\mathbf{Z}$ from the GAE.
\STATE Initialize parameters $\tilde{\boldsymbol{\mu}}$, $\tilde{\boldsymbol{\sigma}}$, and $\boldsymbol{\pi}$ using the fitted GMM.
\FOR{$t = 1$ to $T$}
    \STATE Train \textit{VMGAE} by minimizing $\mathcal{L}_{\text{VMGAE}}(\mathbf{X}, \mathbf{A})$ (Eq.~\eqref{eq:final_loss}).
\ENDFOR
\STATE Fit a final GMM on the learned latent representations $\mathbf{Z}$.
\STATE \textbf{return} Clustering results based on the final GMM.
\end{algorithmic}
\end{algorithm}

\section{Experiments}

\subsection{Experimental Setup and Datasets}
\label{sec:setup}
We employed 19 datasets from the UCR time series classification archive \cite{HUANG20161} for our clustering experiments, with specific details provided in Table \ref{tab:nmi_comparison} and Table \ref{tab:ri_comparison}. Our networks were implemented and tested using PyTorch \cite{paszke2019pytorch}, Torch\_Geometric \cite{fey2019fast}, and executed on an A100 GPU (40G). \textit{VMGAE} was trained with a learning rate of \( 1e^{-4} \) for 500 epochs in full-batch mode, using the Adam optimizer for optimization. Dropout with \( p = 0.01 \) was applied to prevent overfitting. A significant advantage of our method is that we can leverage the latent distribution to tune hyperparameters (as illustrated in Figure \ref{fig:tsne_visualization} in Appendix E) . The hyperparameters \( \gamma \), \( \lambda \), \( W \), and \( \alpha \) were tuned by visualizing the latent distribution of the training set for each dataset separately. During the testing phase, these hyperparameters were fixed, and the final results were evaluated. The details of the datasets used, the sensitivity of hyperparameters, and the evaluation metrics are provided in Appendices C, F.1, and D, respectively.

\subsection{Quantitative Analysis}
The performance of \textit{VMGAE} was benchmarked against several time series clustering methods to evaluate its clustering capabilities thoroughly. The results presented in Tables \ref{tab:nmi_comparison} and \ref{tab:ri_comparison} are sourced from the original papers, except R-Clustering \cite{jorge2024time}, where results were obtained by running the authors' publicly available code. Both tables highlight the best result for each dataset in bold.

As shown in Table \ref{tab:nmi_comparison}, \textit{VMGAE} delivers superior performance, achieving the lowest average rank of 3.1579, the highest average NMI score of 0.6553, and surpassing state-of-the-art (SOTA) methods on seven datasets. Similarly, Table \ref{tab:ri_comparison} highlights \textit{VMGAE}'s strong results based on the Rand Index (RI) metric, with the lowest average rank of 2.6842, the highest average RI of 0.8605, and outperforming SOTA across seven datasets. Notably, on specific datasets such as \textit{TwoPatterns}, \textit{SonyAIBORobotSurface}, and \textit{TwoLeadECG}, the SOTA results were significantly exceeded, with NMI improvements of 0.5089, 0.2212, and 0.1815, and RI improvements of 0.1471, 0.0991, and 0.0169, respectively. 

Further extensive qualitative analysis of our method is provided in Appendix E.

\subsection{Application in Finance}
Understanding stock market dynamics in finance is essential for making informed investment decisions. Detecting patterns and communities within this complex network of stocks helps gain insights into market behavior and make better investment choices.

\begin{figure}[!t]
\hspace{-4.5mm}\resizebox{0.51\textwidth}{!}{ 
\centering
\subfloat[]{\includegraphics[width=2.5in]{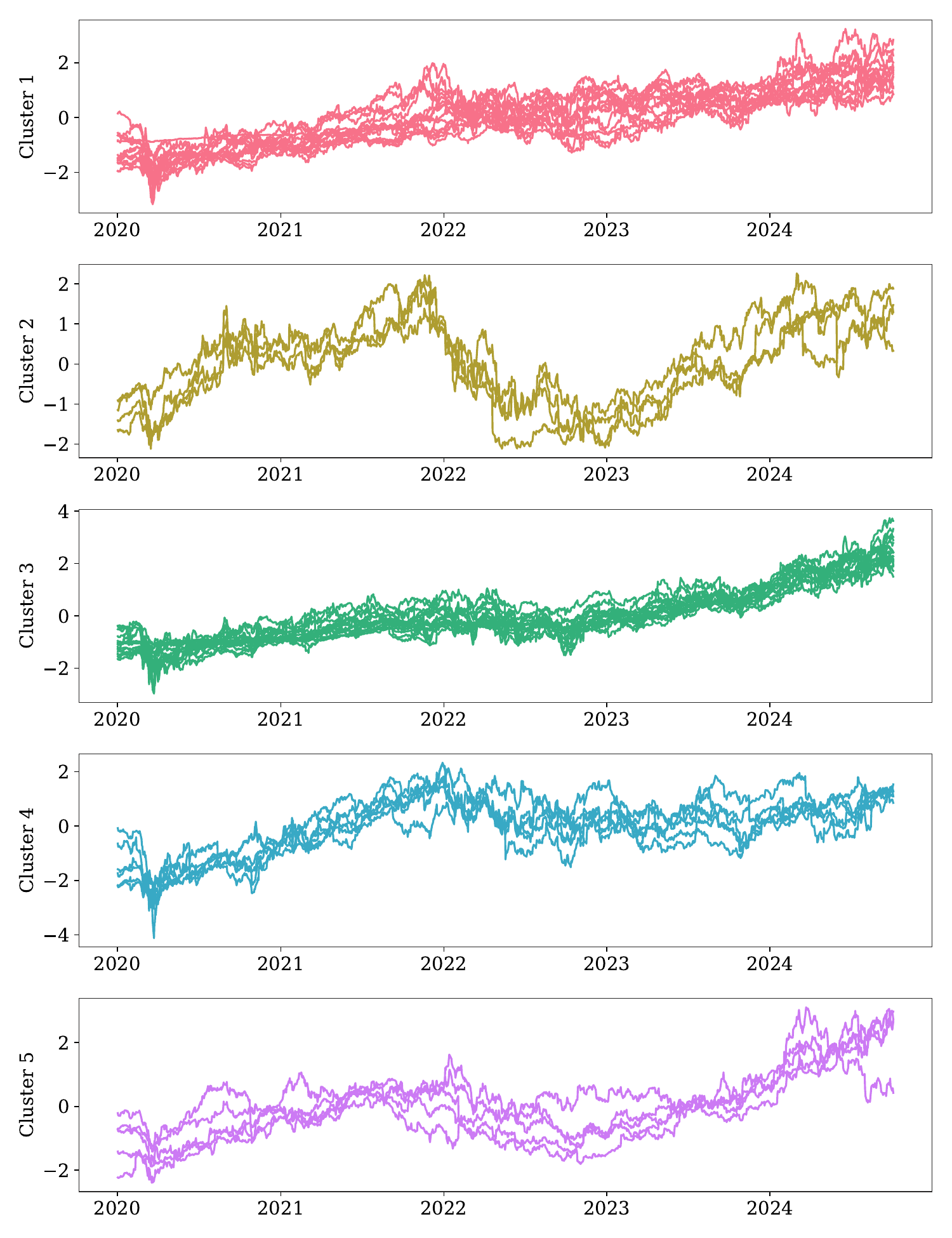}%
\label{fig_first_case1}}
\hfil
\subfloat[]{\includegraphics[width=2.5in]{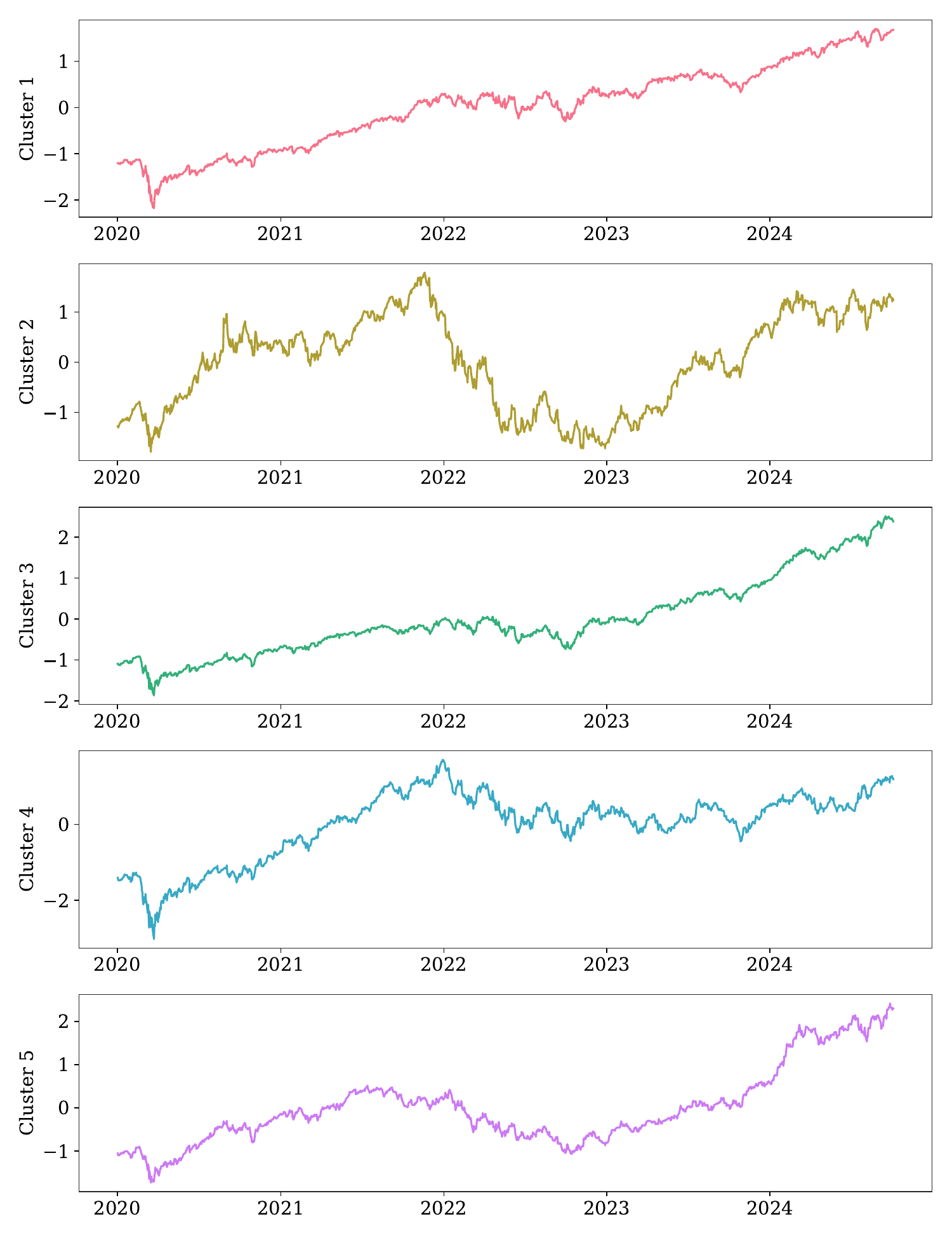}%
\label{fig_second_case2}}
}
\caption{(a) Clustering results of the normalized closing prices for the top 50 U.S. stocks, grouped into five clusters. (b) The average normalized closing price for each cluster shows distinct patterns across the clusters.}
\label{fig:stock-market}
\end{figure}

In this section, we demonstrate the effectiveness of our approach by applying it to real-world stock market data and evaluating the quality of the resulting clusters. We selected the top 50 publicly traded U.S. stocks listed on NASDAQ, NYSE, and NYSE American, ranked by market capitalization. The input time series for our model consists of daily normalized closing prices from January 1, 2020, to October 4, 2024. We set the number of clusters to 5 based on the Elbow Method \cite{thorndike1953elbow}. The results are displayed in Figure \ref{fig_first_case1}, with the average for each cluster shown in Figure \ref{fig_second_case2}, highlighting distinct discriminative patterns.

\section{Conclusion}
In this work, we introduce a novel method for clustering time series data by leveraging graph structures, achieving strong performance across various datasets. Our approach transforms time series data into graph representations using Weighted Dynamic Time Warping, enabling the capture of temporal dependencies and structural relationships. We then apply the proposed Variational Mixture Graph Autoencoder (VMGAE) to generate a Gaussian mixture latent space, improving data separation and clustering accuracy. Extensive experiments demonstrate the effectiveness of our method, including sensitivity analysis on hyperparameters and the evaluation of different convolutional layer architectures. Furthermore, we applied our method to real-world financial data, uncovering community structures in stock markets and showcasing its potential benefits for market prediction, portfolio optimization, and risk management. These findings highlight the versatility and practical applications of VMGAE in addressing time series clustering challenges.

\appendix

\section*{Appendix}

\section{A. Derivation of ELBO for VMGAE}

The Evidence Lower Bound (ELBO) for VMGAE is defined as follows:

\footnotesize
\begin{align}
    \log p({\bf A})&=\log \int_{\mathbf{Z}} \sum_{\textbf{c}} p({\bf A,Z},\textbf{c}) \nonumber\\
    &=\log \int_{\mathbf{Z}} \sum_{\textbf{c}}\left[p({\bf A,Z},\textbf{c})\frac{q({\bf Z},\textbf{c}|{\bf X,A})}{q({\bf Z},\textbf{c}|{\bf X,A})}\right]\nonumber\\
    &\geq  E_{ q({\bf Z},\textbf{c}|{\bf X,A})}[ \log\frac{p({\bf A,Z},\textbf{c})}{q({\bf Z},\textbf{c}|{\bf X,A})}]\nonumber\\
    & =\mathcal{L}_{\textup{ELBO}}({\bf X,A}),\label{eqn:loglikelihood2}
\end{align} 
\normalsize

\noindent where \textbf{X} refers to the feature matrix (or time series matrix in our case), and \textbf{A} represents the adjacency matrix. Jensen's inequality is applied to arrive at this bound.

The expanded form of $\mathcal{L}_{\textup{ELBO}}({\bf X,A})$ using \ref{eq:fac} is given by :

\footnotesize
\begin{align}
    \mathcal{L}_{\textup{ELBO}}({\bf X,A})&=E_{q({\bf Z},\mathbf{c}|{\bf X,A})}[\log p({\bf A,Z},\mathbf{c}) -\log q({\bf Z},\mathbf{c}|{\bf X,A})]\nonumber\\
    &= E_{q({\bf Z},\mathbf{c}|{\bf X,A})}[\underbrace{\log p({\bf A|Z})}_{(\RomanNumeralCaps{1})} +\underbrace{\log p({\bf Z}|\mathbf{c})}_{(\RomanNumeralCaps{2})}+\underbrace{\log p(\mathbf{c})}_{(\RomanNumeralCaps{3})}]\nonumber\\
    &- E_{q({\bf Z},\mathbf{c}|{\bf X,A})}[\underbrace{\log q({\bf Z|X,A})}_{(\RomanNumeralCaps{4})} + \underbrace{\log q(\mathbf{c}|{\bf X,A})}_{(\RomanNumeralCaps{5})}].\label{apeq:ELBO}
\end{align}
\normalsize

\noindent Next, we compute the expectations over the various terms in the ELBO.
\newline
\newline
\noindent Term (\RomanNumeralCaps{1}):

\footnotesize
\begin{align}
 &E_{q({\bf Z},\mathbf{c}|{\bf X,A})}[\log p({\bf A|Z})]=\frac{1}{n^2}\sum_{i=1}^n\sum_{j=1}^n\log p({A_{ij}}|{\bf z_i,z_j})\nonumber\\
&\hspace{4mm}=\frac{1}{n^2}\sum_{i=1}^n\sum_{j=1}^n A_{ij}\log{\hat{A}_{ij}}+(1-A_{ij})\log(1-\hat{A}_{ij}).\nonumber
\end{align}
\normalsize

\noindent Term (\RomanNumeralCaps{2}):
\footnotesize
\begin{align}
&E_{q(\mathbf{Z}, \mathbf{c} \mid \mathbf{X},\mathbf{A})}[\log p(\mathbf{Z}\mid \mathbf{c})] \nonumber\\
&\hspace{2mm} =\sum_{i=1}^n\int_{\mathbf{z_i}} \sum_{c_i=1}^K q(c_i \mid \mathbf{X},\mathbf{A}) q(\mathbf{z_i} \mid \mathbf{X},\mathbf{A}) \log p(\mathbf{z_i} \mid c_i) d \mathbf{z_i}\nonumber\\
&\hspace{2mm} =\sum_{i=1}^n\sum_{c_i=1}^K q(c_i \mid \mathbf{X},\mathbf{A}) \int_{\mathbf{z_i}}  \mathcal{N}\left(\mathbf{z_i} \mid {\boldsymbol{\mu_i}}, {\boldsymbol{\sigma_i}}^2 \mathbf{I}\right) \log \mathcal{N}\left(\mathbf{z_i} \mid \tilde{\boldsymbol{\mu}}_{c_i}, \tilde{\boldsymbol{\sigma}}_{c_i}^2 \mathbf{I}\right) d \mathbf{z_i},
\end{align}
\normalsize

According to appendix B \cite{jiang2016variational}, we have:
\footnotesize
\begin{align}
&E_{q(\mathbf{Z}, \mathbf{c} \mid \mathbf{X},\mathbf{A})}[\log p(\mathbf{Z}\mid \mathbf{c})] \nonumber\\
&= -\sum_{i=1}^n\sum_{c_i=1}^K q(c_i \mid \mathbf{X},\mathbf{A})\bigg[\frac{H}{2} \log (2 \pi) \nonumber\\
& +\frac{1}{2}\left(\sum_{h=1}^H \log \tilde{\boldsymbol{\sigma}}_{c_i h}^2+\sum_{h=1}^H \frac{{\boldsymbol{\sigma}}_h^2}{\tilde{\boldsymbol{\sigma}}_{c_i h}^2}+\sum_{h=1}^H \frac{\left({\boldsymbol{\mu}}_h-\tilde{\boldsymbol{\mu}}_{c_i h}\right)^2}{\tilde{\boldsymbol{\sigma}}_{c_i h}^2}\right)\bigg].
\end{align}
\normalsize

\noindent Term (\RomanNumeralCaps{3}):

\footnotesize
\begin{align}
&E_{q(\mathbf{Z}, \mathbf{c} \mid \mathbf{X},\mathbf{A})}[\log p(\mathbf{c})]=\nonumber\\
&\hspace{2mm} =\sum_{i=1}^n\int_{\mathbf{z_i}} q(\mathbf{z_i} \mid \mathbf{X},\mathbf{A}) \sum_{c_i=1}^K q(c_i \mid \mathbf{X},\mathbf{A}) \log \boldsymbol{\pi}_{c_i} d \mathbf{z_i} \nonumber\\
&\hspace{2mm} =\sum_{i=1}^n\sum_{c_i=1}^K q(c_i \mid \mathbf{X},\mathbf{A}) \log \boldsymbol{\pi}_{c_i}.
\end{align}
\normalsize

\noindent Term (\RomanNumeralCaps{4}):

\footnotesize
\begin{align}
&\hspace{-4mm}E_{q(\mathbf{Z}, \mathbf{c} \mid \mathbf{X},\mathbf{A})}[\log q(\mathbf{Z} \mid \mathbf{X},\mathbf{A})] \nonumber\\
&\hspace{2mm} =\sum_{i=1}^n\int_{\mathbf{z_i}} \sum_{c_i=1}^K q(c_i \mid \mathbf{X},\mathbf{A}) q(\mathbf{z_i} \mid \mathbf{X},\mathbf{A}) \log q(\mathbf{z_i} \mid \mathbf{X},\mathbf{A}) d \mathbf{z_i} \nonumber\\
&\hspace{2mm} =\int_{\mathbf{z}} \mathcal{N}\left(\mathbf{z} ; {\boldsymbol{\mu}}, {\boldsymbol{\sigma}}^2 \mathbf{I}\right) \log \mathcal{N}\left(\mathbf{z} ; {\boldsymbol{\mu}}, {\boldsymbol{\sigma}}^2 \mathbf{I}\right) d \mathbf{z}\nonumber\\
&\hspace{2mm} =-\frac{H}{2} \log (2 \boldsymbol{\pi}) 
 -\frac{1}{2} \sum_{h=1}^H\left(1+\log {\boldsymbol{\sigma}}_h^2\right).
\end{align}
\normalsize

\noindent Term (\RomanNumeralCaps{5}):

\footnotesize
\begin{align}
&\hspace{-4mm}E_{q(\mathbf{Z}, \mathbf{c} \mid \mathbf{X},\mathbf{A})}[\log q(\mathbf{c} \mid \mathbf{X},\mathbf{A})] \nonumber\\
&\hspace{-2mm}=\sum_{i=1}^n\int_{\mathbf{z_i}}\sum_{c_i=1}^K q(\mathbf{z_i} \mid \mathbf{X},\mathbf{A}) q(c_i \mid \mathbf{X},\mathbf{A}) \log q(c_i \mid \mathbf{X},\mathbf{A}) d \mathbf{z_i} \nonumber\\
&\hspace{-2mm} =\sum_{i=1}^n\int_{\mathbf{z_i}} q(\mathbf{z_i} \mid \mathbf{X},\mathbf{A}) \sum_{c_i=1}^K q(c_i \mid \mathbf{X},\mathbf{A}) \log q(c_i \mid \mathbf{X}) d \mathbf{z_i} \nonumber\\
&\hspace{-2mm} =\sum_{i=1}^n\sum_{c_i=1}^K q(c_i \mid \mathbf{X},\mathbf{A}) \log q(c_i \mid \mathbf{X},\mathbf{A}).
\end{align}
\normalsize
\noindent By putting all terms together, we will have:

\footnotesize
\begin{align}
&\mathcal{L}_{\text {ELBO }}(\mathbf{X},\mathbf{A}) =\frac{1}{n^2}\sum_{i=1}^n\sum_{j=1}^n A_{ij}\log{\hat{A}_{ij}} +(1-A_{ij})\log(1-\hat{A}_{ij})\nonumber \\
&\hspace{4mm} -\frac{1}{2}\sum_{i=1}^n \sum_{c_i=1}^K q(c_i \mid \mathbf{X},\mathbf{A}) \sum_{h=1}^H\bigg( \log \tilde{\boldsymbol{\sigma}}_{c_i h}^2+ \frac{{\boldsymbol{\sigma}}_h^2}{\tilde{\boldsymbol{\sigma}}_{c_i h}^2}+\nonumber\\
&\hspace{4mm} \frac{\left({\boldsymbol{\mu}}_h-\tilde{\boldsymbol{\mu}}_{c_i h}\right)^2}{\tilde{\boldsymbol{\sigma}}_{c_i h}^2}\bigg)  +\sum_{i=1}^n\sum_{c_i=1}^K q(c_i \mid \mathbf{X},\mathbf{A}) \log \frac{\boldsymbol{\pi}_{c_i}}{q(c_i \mid \mathbf{X},\mathbf{A})}\nonumber\\ 
&\hspace{4mm} +\frac{1}{2} \sum_{h=1}^H\left(1+\log {\boldsymbol{\sigma}_h^2}\right).
\end{align}
\normalsize

\section{B. Derivation of $q(c_i | \mathbf{X},\mathbf{A})$}
\label{sec:qc}
An important point is how to calculate $q(c_i\mid \mathbf{X},\mathbf{A})$. We can reformat the ELBO into the following form:

\footnotesize
\begin{align}
 &\hspace{-4mm}\mathcal{L}_{\mathrm{ELBO}}(\mathbf{X},\mathbf{A})=E_{q(\mathbf{Z}, \mathbf{c} \mid \mathbf{X},\mathbf{A})}\left[\log \frac{p(\mathbf{A}, \mathbf{Z}, \mathbf{c})}{q(\mathbf{Z}, \mathbf{c} \mid \mathbf{X},\mathbf{A})}\right] \nonumber\\
&\hspace{-2mm} = E_{q(\mathbf{Z}, \mathbf{c} \mid \mathbf{X},\mathbf{A})}\left[\log \frac{p(\mathbf{A}\mid \mathbf{Z})p(\mathbf{c}\mid \mathbf{Z})p(\mathbf{Z})}{q(\mathbf{Z}\mid \mathbf{X},\mathbf{A})q(\mathbf{c}\mid \mathbf{X},\mathbf{A})}\right] \nonumber\\
&\hspace{-2mm} = E_{q(\mathbf{Z}, \mathbf{c} \mid \mathbf{X},\mathbf{A})}\left[\log \frac{p(\mathbf{A}\mid \mathbf{Z})p(\mathbf{Z})}{q(\mathbf{Z}\mid \mathbf{X},\mathbf{A})}+\log \frac{p(\mathbf{c}\mid \mathbf{Z})}{q(\mathbf{c}\mid \mathbf{X},\mathbf{A})} \right]\nonumber\\
&\hspace{-2mm} = E_{q(\mathbf{Z} \mid \mathbf{X},\mathbf{A})}\left[\log \frac{p(\mathbf{A}\mid \mathbf{Z})p(\mathbf{Z})}{q(\mathbf{Z}\mid \mathbf{X},\mathbf{A})}\right] \nonumber\\
&-\sum_{i=1}^n\int_{\mathbf{z_i}} q(\mathbf{z_i} \mid \mathbf{X},\mathbf{A}) \sum_{c_i} q(c_i \mid \mathbf{X},\mathbf{A}) \log \frac{q(c_i \mid \mathbf{X},\mathbf{A})}{p(c_i \mid \mathbf{z_i})} d \mathbf{z_i} \nonumber\\
&\hspace{-2mm} = E_{q(\mathbf{Z} \mid \mathbf{X},\mathbf{A})}\left[\log \frac{p(\mathbf{A}\mid \mathbf{Z})p(\mathbf{Z})}{q(\mathbf{Z}\mid \mathbf{X},\mathbf{A})}\right]\nonumber\\
& -\sum_{i=1}^n\int_{\mathbf{z_i}} q(\mathbf{z_i} \mid \mathbf{X},\mathbf{A}) \KL(q(c_i \mid \mathbf{X},\mathbf{A}) \| p(c_i \mid \mathbf{z_i})) d \mathbf{z_i}.
\end{align}
\normalsize

\noindent In the Equation above, the first term is not dependent on $\mathbf{c}$ and the second is non-negative. Hence, to maximize $\mathcal{L}_{\text {ELBO }}(\mathbf{X},\mathbf{A}), D_{K L}(q(c_i \mid \mathbf{X},\mathbf{A}) \| p(c_i \mid \mathbf{z_i})) \equiv 0$ should be satisfied. As a result, we use the following Equation to compute $q(c_i \mid \mathbf{X},\mathbf{A})$ in \textit{VMGAE}:

$$
q(c_i \mid \mathbf{X},\mathbf{A})=p(c_i \mid \mathbf{z_i}) \equiv \frac{p(c_i) p(\mathbf{z_i} \mid c_i)}{\sum_{c^{\prime}=1}^K p\left(c^{\prime}\right) p\left(\mathbf{z_i} \mid c^{\prime}\right)}.
$$

\section{C. Datasets}\label{sec:Datasets}

We conducted our clustering experiments using 19 datasets from the UCR Time Series Classification Archive \cite{HUANG20161}, a widely recognized benchmark for time series analysis. The details of these datasets are presented in Table \ref{tab:datasets}.

\begin{table}[htbp]
    \centering
    \resizebox{0.45\textwidth}{!}{
    \begin{tabular}{lccccc}
        \toprule
        No. & Name & Train/Test & Length & Classes \\
        \midrule
        1 & Beef & 30/30 & 470 & 5\\
        2 & Car & 60/60 & 577 & 4\\
        3 & DiatomSizeReduction & 16/306 & 345 & 4\\
        4 & Dist.Phal.Outl.AgeGroup & 400/139 & 80 & 3\\
        5 & ECG200 & 100/100 & 96 & 2\\
        6 & ECGFiveDays & 23/861 & 136 & 2\\
        7 & Meat & 60/60 & 448 & 3\\
        8 & Mid.Phal.TW & 399/154 & 80 & 6\\
        9 & OSULeaf & 200/242 & 427 & 6\\
        10 & Plane & 105/105 & 144 & 7\\
        11 & Prox.Phal.Outl.AgeGroup & 400/205 & 80 & 3\\
        12 & SonyAIBORobotSurface & 20/601 & 70 & 2\\
        13 & SwedishLeaf & 500/625 & 128 & 15\\
        14 & Symbols & 25/995 & 398 & 6\\
        15 & ToeSegmentation1 & 40/228 & 277 & 2\\
        16 & TwoPatterns & 1000/4000 & 128 & 4\\
        17 & TwoLeadECG & 23/1139 & 82 & 2\\
        18 & Wafer & 1000/6164 & 152 & 2\\
        19 & WordSynonyms & 267/638 & 270 & 25\\
        \bottomrule
    \end{tabular}
    }
    \caption{Statistics of the 19 datasets from the UCR benchmark used in our experiments.}
    \label{tab:datasets}
\end{table}

\section{D. Evaluation Metrics} \label{sec:eval_metrics}
We evaluate the clustering performance in our analysis using two well-established metrics: the Rand Index (\(RI\)) and Normalized Mutual Information (\(NMI\)). The Rand Index, which quantifies the agreement between the predicted and actual clustering assignments, is computed as follows:

\footnotesize
\begin{equation}
\label{eq:ri}
RI = \frac{TP + TN}{TP + FP + FN + TN}.
\end{equation}
\normalsize

In this expression, \(TP\) (True Positive) denotes the number of pairs of time series correctly classified into the same cluster, while \(TN\) (True Negative) refers to the number of pairs accurately assigned to different clusters. Conversely, \(FP\) (False Positive) captures the number of pairs incorrectly grouped into the same cluster, and \(FN\) (False Negative) accounts for pairs that should be clustered together but are mistakenly separated.

The \(NMI\) score is defined as:

\footnotesize
\begin{equation}
\label{eq:nmi}
NMI = \frac{\sum_{i=1}^{K} \sum_{j=1}^{K} N_{ij} \log \left(\frac{n \cdot N_{ij}}{|G_i| \cdot |P_j|}\right)}{\sqrt{\left(\sum_{i=1}^{K} |G_i| \log \left(\frac{|G_i|}{n}\right)\right) \cdot \left(\sum_{j=1}^{M} |P_j| \log \left(\frac{|P_j|}{n}\right)\right)}},
\end{equation}
\normalsize

\noindent where \(N_{ij}\) represents the number of time series that are common between the \(i\)-th ground truth cluster \(G_i\) and the \(j\)-th predicted cluster \(P_j\). \(|\cdot|\) is the number of time series in the cluster. The variables $K$ and $n$ in Equations \ref{eq:ri} and \ref{eq:nmi} are defined as previously explained in the section Notation.

\section{E. Qualitative Analysis} 
\label{sec:Qualitative_Analysis}

\begin{figure*}[!t]
\centering
\subfloat[]{\includegraphics[width=2.2in]{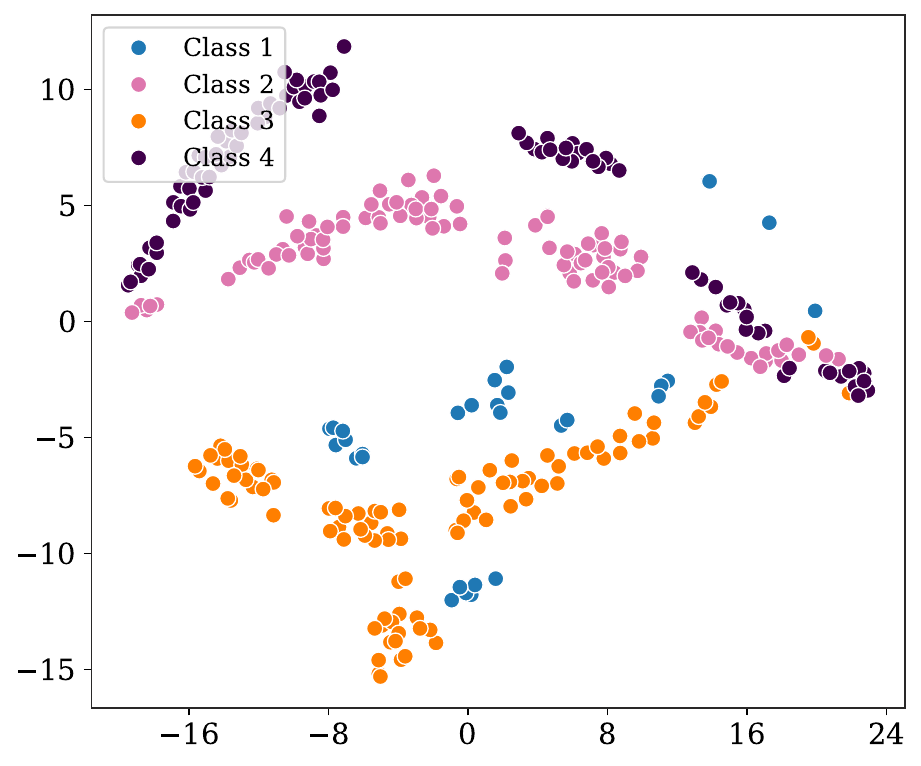}%
\label{fig_1}}
\subfloat[]{\includegraphics[width=2.2in]{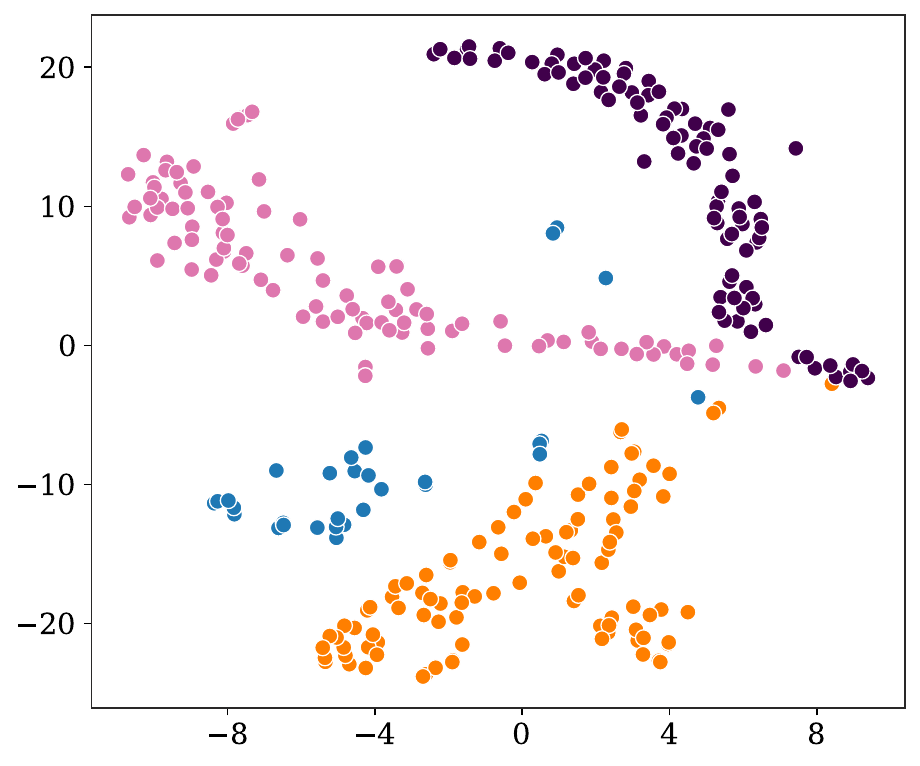}%
\label{fig_2}}
\subfloat[]{\includegraphics[width=2.2in]{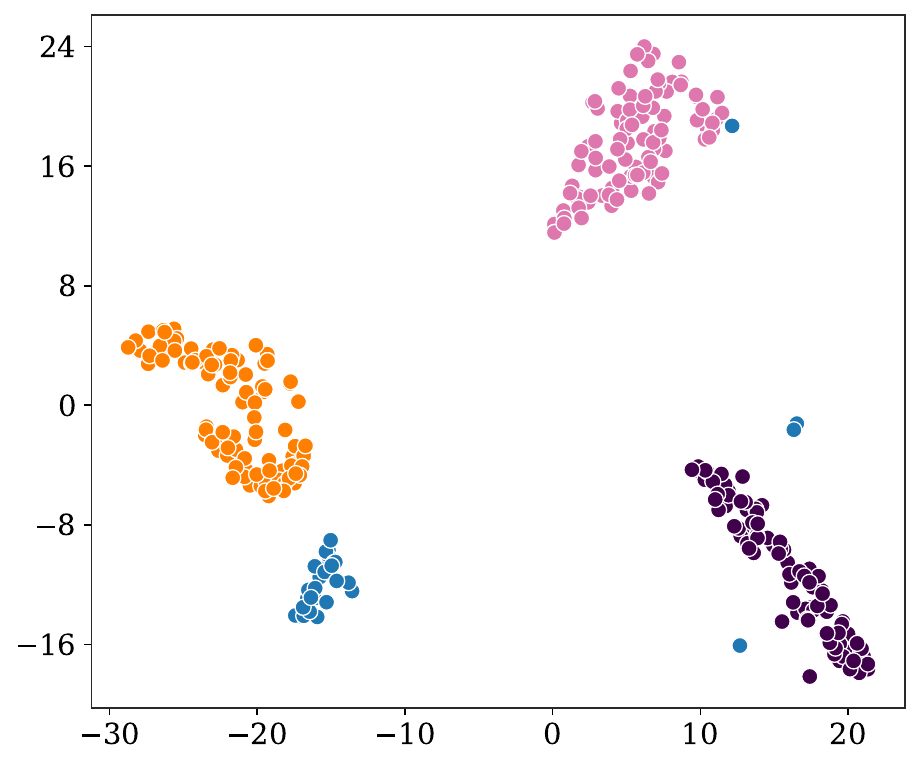}%
\label{fig_3}}
    \caption{The visualizations with t-SNE on the dataset \textit{DiatomSizeReduction}. The colors of the points indicate the actual labels. (a) epoch 0, (b) epoch 10, (c) epoch 100.}
\label{fig:tsne_visualization}
\end{figure*}

We further present visualizations of the evolving clusters during training on the DiatomSizeReduction in Figure \ref{fig:tsne_visualization}. These clusters are mapped from the latent space representations \( \mathbf{Z} \) to a 2D space using t-SNE \cite{JMLR:v9:vandermaaten08a}. The t-SNE plots illustrate how the latent representations become increasingly well-separated as training progresses, reflecting \textit{VMGAE}'s capacity to learn distinct clusters from the time series data. 

\section{F. Ablation Study}

\subsection{F.1. Hyperparameter Sensitivity Analysis}
In this section, we analyze the impact and sensitivity of the hyperparameters \( \gamma \), \( \lambda \), and \( \alpha \) on our method. To assess the sensitivity of each hyperparameter, the other hyperparameters were kept fixed at their optimal values, as shown in Table \ref{tab:hyper_parameter_impact}. The hyperparameter values \( \gamma = 0.7 \) and \( \gamma = 1.0 \) yield better metric results for the \textit{SonyAIBORobotSurface1} dataset compared to \( \gamma = 0.2 \), which was used to report the results in Tables \ref{tab:nmi_comparison} and \ref{tab:ri_comparison}. This improvement was not evident through the visualization process. As shown in the table, for some datasets like \textit{Meat}, the model is not sensitive to the hyperparameter values, whereas for other datasets, such as \textit{Car}, the model shows some sensitivity to the hyperparameter values.

\subsection{F.2. Impact of Convolutional Layer Variants}
\label{sec:convolution_layers_performance}

Several advanced graph convolutional layers have been developed to enhance information propagation in graph neural networks, each with distinct methods and advantages. One well-known type of convolutional layer is the Graph Attention Network (GAT) \cite{veličković2018graphattentionnetworks}. GAT layers introduce attention mechanisms to graph convolutions, enabling the model to assign different importance to neighboring nodes rather than treating them uniformly. Specifically, the GAT layer computes attention coefficients \( \alpha_{ij} \) based on node features, which are then used to aggregate information from neighboring nodes. The process of each GAT layer is expressed as follows:
\begin{equation}
\mathbf{Z}_i^{(l+1)} = \phi\left(\sum_{j \in \mathcal{N}(i)} \alpha_{ij} \mathbf{W}^{(l)} \mathbf{Z}_j^{(l)}\right),
\end{equation}
where \( \mathcal{N}(i) \) denotes the neighbors of node \( i \), and \( \phi \) is an activation function. The attention mechanism allows GAT layers to dynamically adjust the influence of neighboring nodes, leading to more flexible and potentially more accurate embeddings.

Another variant is SAGEConv \cite{hamilton2018inductiverepresentationlearninglarge}, which stands for Sample and Aggregation Convolution. This layer generalizes GCNs by allowing for aggregating features from a sampled set of neighbors instead of using all neighbors. Various aggregation operators like mean aggregator, LSTM aggregator, and polling aggregator can perform the aggregation process. The final formula is given by :

\footnotesize
\begin{align}
&\mathbf{Z}_i^{(l+1)} = \phi \bigg(\mathbf{W}_1^{(l)} \mathbf{Z}_i^{(l)} + \mathbf{W}_2^{(l)}\text{AGGREGATE}(\{\mathbf{Z}_j^{(l)}: j \in \mathcal{N}(i)\})\bigg),
\end{align}
\normalsize 

where \text{AGGREGATE} is a function that combines the features of the neighbors.

ChebConv \cite{defferrard2017convolutionalneuralnetworksgraphs} is another robust convolutional layer that utilizes a recursive process to produce $ Z_i^{j}$'s and aggregate them by some learnable parameters. The ChebConv whole operation is given by:

\footnotesize
\begin{align}
&\mathbf{Z_i^{(1)} = X_i}\nonumber\\ 
&\mathbf{Z_i^{(2)} = \tilde{L}.X_i}\nonumber\\  
&\mathbf{Z_i^{(k)} = 2 \tilde{L}.Z_i^{k-1}-Z_i^{k-2}}\nonumber\\
&\mathbf{\tilde{X_i}} = \sum_{k=0}^{K} \mathbf{\Theta _k}\mathbf{Z_i^k},
\end{align}
\normalsize

\noindent where \( \mathbf{T}_k(\mathbf{L}) \) denotes the Chebyshev polynomial of order \( k \), and \( \mathbf{L} \) is the graph Laplacian.

\begin{table*}[t]
\begin{center}
\resizebox{0.9\textwidth}{!}{ 
\begin{minipage}{\textwidth}

\begin{tabular}{@{}ll|cccc|cccc|cccc@{}}
\toprule
     & Dataset    & \multicolumn{4}{c|}{Car} & \multicolumn{4}{c|}{Meat} & \multicolumn{4}{c}{SonyAIBORobotSurface1} \\ \midrule
            & $\alpha$   & 0.025   & 0.05  & 0.075   & 0.1  & 0.025   & 0.05  & 0.075   & 0.1 & 0.025  & 0.05  & 0.075   & 0.1 \\ \midrule
  &  \multicolumn{1}{c|}{NMI}  & 0.4906 & 0.5634 & \textbf{0.6193}  & 0.4884 & \textbf{1.0000} & \textbf{1.0000}  & \textbf{1.0000}  & 0.8996 & 0.8524 & 0.9182  & \textbf{0.9319} & 0.9089  \\
    &  \multicolumn{1}{c|}{RI}  & 0.7322 & 0.7813 & \textbf{0.8045}  & 0.7559  & \textbf{1.0000} & \textbf{1.0000} & \textbf{1.0000}  & 0.9570  & 0.9544 & 0.9801 & \textbf{0.9834}  & 0.9769  \\
\bottomrule
\end{tabular}
\par\centering (a) Impact of hyperparameter $\alpha$

\begin{tabular}{@{}ll|cccc|cccc|cccc@{}}
\toprule
     & Dataset    & \multicolumn{4}{c|}{Car} & \multicolumn{4}{c|}{Meat} & \multicolumn{4}{c}{SonyAIBORobotSurface1} \\ \midrule
            & $\lambda$   & 0.1   & 0.01  & 0.001   & 0.0001  & 0.1   & 0.01  & 0.001   & 0.0001 & 0.1  & 0.01  & 0.001   & 0.0001 \\ \midrule
  &  \multicolumn{1}{c|}{NMI}  & 0.4421 & 0.6088 & \textbf{0.6193}  & \textbf{0.6193}  & \textbf{1.0000} & \textbf{1.0000} & \textbf{1.0000}  & \textbf{1.0000}  & 0.9298 & \textbf{0.9319} & 0.9298  & 0.9298  \\
    &  \multicolumn{1}{c|}{RI}  & 0.7384 & 0.7920 & \textbf{0.8045}  & \textbf{0.8045}  & \textbf{1.0000} & \textbf{1.0000} & \textbf{1.0000}  & \textbf{1.0000}  & \textbf{0.9834} & \textbf{0.9834} & \textbf{0.9834}  & \textbf{0.9834}  \\
\bottomrule
\end{tabular}
\par\centering (b) Impact of hyperparameter $\lambda$

\begin{tabular}{@{}ll|cccc|cccc|cccc@{}}
\toprule
     & Dataset    & \multicolumn{4}{c|}{Car} & \multicolumn{4}{c|}{Meat} & \multicolumn{4}{c}{SonyAIBORobotSurface1} \\ \midrule
            & $\gamma$   & 0.2   & 0.4  & 0.7   & 1.0  & 0.2   & 0.4  & 0.7 & 1.0 & 0.2   & 0.4  & 0.7 & 1.0 \\ \midrule
  &  \multicolumn{1}{c|}{NMI}  & 0.5308 & \textbf{0.6193} & 0.5483  & 0.4895  & \textbf{1.0000} & \textbf{1.0000} & \textbf{1.0000}  & \textbf{1.0000}& 0.9319 & 0.9190 & \textbf{0.9427}  & \textbf{0.9427}  \\
    &  \multicolumn{1}{c|}{RI}  & 0.7751 & \textbf{0.8045} & 0.7853  & 0.7661  & \textbf{1.0000} & \textbf{1.0000} & \textbf{1.0000}  & \textbf{1.0000} & 0.9834 & 0.9801 & \textbf{0.9867}  & \textbf{0.9867}  \\
\bottomrule
\end{tabular}
\par\centering (c) Impact of hyperparameter $\gamma$
\caption{The tables show the impact of different hyperparameters on the NMI and RI metrics for 3 different datasets.}
\label{tab:hyper_parameter_impact}
\end{minipage}
} 
\end{center}
\end{table*}

Similarly, SGConv \cite{wu2019simplifying}, or Simplifying Graph Convolution, provides an efficient alternative that simplifies the graph convolution operation while maintaining good performance. The operation can be expressed as:

\footnotesize
\begin{align}
\mathbf{Z_i} = \text{Softmax}\mathbf{\left(S^kx_i\Theta\right)},
\end{align}
\normalsize

\noindent where $S$ is the normalized adjacency matrix and $k$ is a fixed number and $\Theta$ is the laearnable parameter matrix.

Finally, TAGConv \cite{du2018topologyadaptivegraphconvolutional}, or Adaptive Graph Convolution, adapts the convolution operation based on the local graph structure. It computes the convolution by taking into account the varying degrees of nodes:
\begin{equation}
\mathbf{Z}_i = \sum_{k=0}^{K} \mathbf{A}^{k} \mathbf{X}_iW_k,
\end{equation}
where \(A\) is the normalized adjacency matrix and \(W\)'s are learnable parameters.


we examines how different convolutional layers affect the model's ability to learn node embeddings and perform clustering. In the main results shown in Tables \ref{tab:nmi_comparison} and \ref{tab:ri_comparison}, we used Graph Convolutional Network (GCN) layers. Here, we test other types of convolutional layers and compare their effects on the model's performance across different datasets. The results of these comparisons are shown in Table \ref{tab:conv_comparison}. 

\begin{table*}[ht]
\centering

\begin{adjustbox}{max width=\textwidth}
\begin{tabular}{
>{\centering\arraybackslash}p{2.5cm} 
>{\centering\arraybackslash}p{1.2cm} >{\centering\arraybackslash}p{1.2cm} 
>{\centering\arraybackslash}p{1.2cm} >{\centering\arraybackslash}p{1.2cm} 
>{\centering\arraybackslash}p{1.2cm} >{\centering\arraybackslash}p{1.2cm}}
\toprule
{Conv. Layer} & 
\multicolumn{2}{c}{{Beef}} & 
\multicolumn{2}{c}{{Car}} & 
\multicolumn{2}{c}{{Dist. Age Group}} \\

\cmidrule(lr){2-3} \cmidrule(lr){4-5} \cmidrule(lr){6-7}
 & {NMI} & {RI} & {NMI} & {RI} & {NMI} & {RI} \\ 
\midrule
GCN       & \textbf{0.5237} & \textbf{0.7862} & \textbf{0.6193} & \textbf{0.8045} & 0.4400 & 0.6827 \\
GAT       & 0.4926 & 0.7463 & 0.4165 & 0.6548 & 0.4322 & 0.7426 \\
SAGEConv  & 0.4907 & 0.7429 & 0.4499 & 0.7119 & \textbf{0.4637} & \textbf{0.7492} \\
ChebConv  & 0.2789 & 0.7152 & 0.1683 & 0.6534 & 0.3243 & 0.5933 \\
SGConv    & 0.4673 & 0.7418 & 0.4900 & 0.7402 & 0.4363 & 0.7405 \\
TAGConv   & 0.4907 & 0.7429 & 0.4304 & 0.7122 & 0.3876 & 0.7218 \\
\bottomrule
\end{tabular}
\end{adjustbox}
\caption{Performance comparison of different convolutional layers on clustering across datasets (Beef, Car, Distinct Age Group), evaluated using Normalized Mutual Information (NMI) and Rand Index (RI).}
\label{tab:conv_comparison}
\end{table*}

\subsection{F.3. Versatility of VMGAE: Application to Graph Datasets}
While our primary contribution focuses on applying VMGAE to time series data transformed into graph representations, it is important to highlight the versatility of our method, which can be effectively applied to any graph input. The architecture is designed to learn meaningful latent representations across diverse graph datasets.

To demonstrate this, we employed the \textit{Cora} dataset, a benchmark graph dataset comprising scientific publications grouped into distinct categories, with citation relationships forming the edges between nodes. Each node corresponds to a publication, and the edges represent citation links. This dataset is commonly used in graph-based machine-learning tasks due to its structured graph topology and rich node features.

Our experiments on the \textit{Cora} dataset further validate the flexibility of our VMGAE architecture. For this evaluation, the learning rate was set to $1e^{-5}$ the $\lambda$ parameter was set to 0.001, the model was trained for 500 epochs, and dropout was applied with a rate of 0.01. Table \ref{tab:nmi_cora} provides a comparison of NMI scores between VMGAE and other recognized graph-based methods such as GAE, VGAE \cite{kipf2016variational}, and ARGA \cite{pan2019adversariallyregularizedgraphautoencoder}.

\begin{table}[htbp]
    \centering
    \adjustbox{max width=\textwidth}{
    \begin{tabular}{lcccc}
        \toprule
        & VMGAE & ARGA & VGAE & GAE \\
        \midrule
        NMI & \textbf{0.459} & 0.450 & 0.436 & 0.429 \\
        \bottomrule
    \end{tabular}
    }
    \caption{NMI Comparisons on cora data-set}
    \label{tab:nmi_cora}
\end{table}

\bibliography{aaai25}

\begin{thebibliography}{56}
\providecommand{\natexlab}[1]{#1}

\bibitem[{Aghabozorgi, Shirkhorshidi, and Wah(2015)}]{aghabozorgi2015time}
Aghabozorgi, S.; Shirkhorshidi, A.~S.; and Wah, T.~Y. 2015.
\newblock Time-series clustering--a decade review.
\newblock \emph{Information systems}, 53: 16--38.

\bibitem[{Babu, Geethanjali, and Satyanarayana(2012)}]{babu2012clustering}
Babu, M.~S.; Geethanjali, N.; and Satyanarayana, B. 2012.
\newblock Clustering approach to stock market prediction.
\newblock \emph{International Journal of Advanced Networking and Applications}, 3(4): 1281.

\bibitem[{Cao et~al.(2020)Cao, Wang, Duan, Zhang, Zhu, Huang, Tong, Xu, Bai, Tong et~al.}]{cao2020spectral}
Cao, D.; Wang, Y.; Duan, J.; Zhang, C.; Zhu, X.; Huang, C.; Tong, Y.; Xu, B.; Bai, J.; Tong, J.; et~al. 2020.
\newblock Spectral temporal graph neural network for multivariate time-series forecasting.
\newblock \emph{Advances in neural information processing systems}, 33: 17766--17778.

\bibitem[{Chaudhuri and Ghosh(2016)}]{chaudhuri2016using}
Chaudhuri, T.~D.; and Ghosh, I. 2016.
\newblock Using clustering method to understand Indian stock market volatility.
\newblock \emph{arXiv preprint arXiv:1604.05015}.

\bibitem[{Close and Kashef(2020)}]{close2020combining}
Close, L.; and Kashef, R. 2020.
\newblock Combining artificial immune system and clustering analysis: A stock market anomaly detection model.
\newblock \emph{Journal of Intelligent Learning Systems and Applications}, 12(04): 83--108.

\bibitem[{Defferrard, Bresson, and Vandergheynst(2017)}]{defferrard2017convolutionalneuralnetworksgraphs}
Defferrard, M.; Bresson, X.; and Vandergheynst, P. 2017.
\newblock Convolutional Neural Networks on Graphs with Fast Localized Spectral Filtering.
\newblock arXiv:1606.09375.

\bibitem[{Deng and Hooi(2021)}]{deng2021graph}
Deng, A.; and Hooi, B. 2021.
\newblock Graph neural network-based anomaly detection in multivariate time series.
\newblock In \emph{Proceedings of the AAAI conference on artificial intelligence}, volume~35, 4027--4035.

\bibitem[{Ding, Chen, and Bressler(2006)}]{ding2006granger}
Ding, M.; Chen, Y.; and Bressler, S.~L. 2006.
\newblock Granger causality: basic theory and application to neuroscience.
\newblock \emph{Handbook of time series analysis: recent theoretical developments and applications}, 437--460.

\bibitem[{Du et~al.(2018)Du, Zhang, Wu, Moura, and Kar}]{du2018topologyadaptivegraphconvolutional}
Du, J.; Zhang, S.; Wu, G.; Moura, J. M.~F.; and Kar, S. 2018.
\newblock Topology Adaptive Graph Convolutional Networks.
\newblock arXiv:1710.10370.

\bibitem[{Fang, Xu, and Jiang(2020)}]{fang2020survey}
Fang, Y.; Xu, H.; and Jiang, J. 2020.
\newblock A survey of time series data visualization research.
\newblock In \emph{IOP Conference Series: Materials Science and Engineering}, volume 782, 022013. IOP Publishing.

\bibitem[{Fey and Lenssen(2019)}]{fey2019fast}
Fey, M.; and Lenssen, J.~E. 2019.
\newblock Fast graph representation learning with PyTorch Geometric.
\newblock In \emph{ICLR Workshop on Representation Learning on Graphs and Manifolds}.

\bibitem[{Guo et~al.(2017)Guo, Liu, Zhu, and Yin}]{guo2017deep}
Guo, X.; Liu, X.; Zhu, E.; and Yin, J. 2017.
\newblock Deep clustering with convolutional autoencoders.
\newblock In \emph{Neural Information Processing: 24th International Conference, ICONIP 2017, Guangzhou, China, November 14-18, 2017, Proceedings, Part II 24}, 373--382. Springer.

\bibitem[{Hamilton, Ying, and Leskovec(2018)}]{hamilton2018inductiverepresentationlearninglarge}
Hamilton, W.~L.; Ying, R.; and Leskovec, J. 2018.
\newblock Inductive Representation Learning on Large Graphs.
\newblock arXiv:1706.02216.

\bibitem[{Han and Woo(2022)}]{han2022learning}
Han, S.; and Woo, S.~S. 2022.
\newblock Learning sparse latent graph representations for anomaly detection in multivariate time series.
\newblock In \emph{Proceedings of the 28th ACM SIGKDD Conference on knowledge discovery and data mining}, 2977--2986.

\bibitem[{Hird and McDermid(2009)}]{hird2009noise}
Hird, J.~N.; and McDermid, G.~J. 2009.
\newblock Noise reduction of NDVI time series: An empirical comparison of selected techniques.
\newblock \emph{Remote Sensing of Environment}, 113(1): 248--258.

\bibitem[{Huang et~al.(2016)Huang, Ye, Xiong, Lau, Jiang, and Wang}]{HUANG20161}
Huang, X.; Ye, Y.; Xiong, L.; Lau, R.~Y.; Jiang, N.; and Wang, S. 2016.
\newblock Time series k-means: A new k-means type smooth subspace clustering for time series data.
\newblock \emph{Information Sciences}, 367-368: 1--13.

\bibitem[{Ismail~Fawaz et~al.(2019)Ismail~Fawaz, Forestier, Weber, Idoumghar, and Muller}]{ismail2019deep}
Ismail~Fawaz, H.; Forestier, G.; Weber, J.; Idoumghar, L.; and Muller, P.-A. 2019.
\newblock Deep learning for time series classification: a review.
\newblock \emph{Data mining and knowledge discovery}, 33(4): 917--963.

\bibitem[{Jiang et~al.(2016)Jiang, Zheng, Tan, Tang, and Zhou}]{jiang2016variational}
Jiang, Z.; Zheng, Y.; Tan, H.; Tang, B.; and Zhou, H. 2016.
\newblock Variational deep embedding: An unsupervised and generative approach to clustering.
\newblock \emph{arXiv preprint arXiv:1611.05148}.

\bibitem[{Jorge and Rub{\'e}n(2024)}]{jorge2024time}
Jorge, M.-B.; and Rub{\'e}n, C. 2024.
\newblock Time series clustering with random convolutional kernels.
\newblock \emph{Data Mining and Knowledge Discovery}, 1--27.

\bibitem[{Kipf and Welling(2016{\natexlab{a}})}]{kipf2016semi}
Kipf, T.~N.; and Welling, M. 2016{\natexlab{a}}.
\newblock Semi-supervised classification with graph convolutional networks.
\newblock \emph{arXiv preprint arXiv:1609.02907}.

\bibitem[{Kipf and Welling(2016{\natexlab{b}})}]{kipf2016variational}
Kipf, T.~N.; and Welling, M. 2016{\natexlab{b}}.
\newblock Variational graph auto-encoders.
\newblock \emph{arXiv preprint arXiv:1611.07308}.

\bibitem[{Lee, Kim, and Sim(2024)}]{lee2024temporal}
Lee, J.; Kim, D.; and Sim, S. 2024.
\newblock Temporal Multi-features Representation Learning-Based Clustering for Time-Series Data.
\newblock \emph{IEEE Access}.

\bibitem[{Li et~al.(2022)Li, Choi, Xu, Bhowmick, Mah, and Wong}]{li2022autoshape}
Li, G.; Choi, B.; Xu, J.; Bhowmick, S.~S.; Mah, D. N.-y.; and Wong, G. L.-H. 2022.
\newblock Autoshape: An autoencoder-shapelet approach for time series clustering.
\newblock \emph{arXiv preprint arXiv:2208.04313}.

\bibitem[{Li et~al.(2020)Li, Liu, Yang, Liu, Wu, and Wan}]{li2020adaptively}
Li, H.; Liu, J.; Yang, Z.; Liu, R.~W.; Wu, K.; and Wan, Y. 2020.
\newblock Adaptively constrained dynamic time warping for time series classification and clustering.
\newblock \emph{Information Sciences}, 534: 97--116.

\bibitem[{Li, Boubrahimi, and Hamdi(2021)}]{li2021graph}
Li, P.; Boubrahimi, S.~F.; and Hamdi, S.~M. 2021.
\newblock Graph-based clustering for time series data.
\newblock In \emph{2021 IEEE International Conference on Big Data (Big Data)}, 4464--4467. IEEE.

\bibitem[{Liao(2005)}]{liao2005clustering}
Liao, T.~W. 2005.
\newblock Clustering of time series data—a survey.
\newblock \emph{Pattern recognition}, 38(11): 1857--1874.

\bibitem[{Lin et~al.(2012)Lin, Williamson, Borne, and DeBarr}]{lin2012pattern}
Lin, J.; Williamson, S.; Borne, K.; and DeBarr, D. 2012.
\newblock Pattern recognition in time series.
\newblock \emph{Advances in machine learning and data mining for astronomy}, 1(617-645): 3.

\bibitem[{Ma et~al.(2020)Ma, Li, Zhuang, Wang, and Zeng}]{ma2020self}
Ma, Q.; Li, S.; Zhuang, W.; Wang, J.; and Zeng, D. 2020.
\newblock Self-supervised time series clustering with model-based dynamics.
\newblock \emph{IEEE Transactions on Neural Networks and Learning Systems}, 32(9): 3942--3955.

\bibitem[{Ma et~al.(2019)Ma, Zheng, Li, and Cottrell}]{ma2019learning}
Ma, Q.; Zheng, J.; Li, S.; and Cottrell, G.~W. 2019.
\newblock Learning representations for time series clustering.
\newblock \emph{Advances in neural information processing systems}, 32.

\bibitem[{Mudelsee(2019)}]{mudelsee2019trend}
Mudelsee, M. 2019.
\newblock Trend analysis of climate time series: A review of methods.
\newblock \emph{Earth-science reviews}, 190: 310--322.

\bibitem[{Olive et~al.(2020)Olive, Basora, Viry, and Alligier}]{olive2020deep}
Olive, X.; Basora, L.; Viry, B.; and Alligier, R. 2020.
\newblock Deep trajectory clustering with autoencoders.
\newblock In \emph{ICRAT 2020, 9th International Conference for Research in Air Transportation}.

\bibitem[{Pan et~al.(2019)Pan, Hu, Long, Jiang, Yao, and Zhang}]{pan2019adversariallyregularizedgraphautoencoder}
Pan, S.; Hu, R.; Long, G.; Jiang, J.; Yao, L.; and Zhang, C. 2019.
\newblock Adversarially Regularized Graph Autoencoder for Graph Embedding.
\newblock arXiv:1802.04407.

\bibitem[{Paparrizos and Gravano(2015)}]{paparrizos2015k}
Paparrizos, J.; and Gravano, L. 2015.
\newblock k-shape: Efficient and accurate clustering of time series.
\newblock In \emph{Proceedings of the 2015 ACM SIGMOD international conference on management of data}, 1855--1870.

\bibitem[{Paszke et~al.(2019)Paszke, Gross, Massa, Lerer, Bradbury, Chanan, Killeen, Lin, Gimelshein, Antiga et~al.}]{paszke2019pytorch}
Paszke, A.; Gross, S.; Massa, F.; Lerer, A.; Bradbury, J.; Chanan, G.; Killeen, T.; Lin, Z.; Gimelshein, N.; Antiga, L.; et~al. 2019.
\newblock PyTorch: An imperative style, high-performance deep learning library.
\newblock \emph{Advances in Neural Information Processing Systems}, 32: 8024--8035.

\bibitem[{Sakoe(1978)}]{Sakoe1978DynamicPA}
Sakoe, H. 1978.
\newblock Dynamic programming algorithm optimization for spoken word recognition.
\newblock \emph{IEEE Transactions on Acoustics, Speech, and Signal Processing}, 26: 159--165.

\bibitem[{Shah, Isah, and Zulkernine(2019)}]{shah2019stock}
Shah, D.; Isah, H.; and Zulkernine, F. 2019.
\newblock Stock market analysis: A review and taxonomy of prediction techniques.
\newblock \emph{International Journal of Financial Studies}, 7(2): 26.

\bibitem[{Shaukat et~al.(2021)Shaukat, Alam, Luo, Shabbir, Hameed, Li, Abbas, and Javed}]{shaukat2021review}
Shaukat, K.; Alam, T.~M.; Luo, S.; Shabbir, S.; Hameed, I.~A.; Li, J.; Abbas, S.~K.; and Javed, U. 2021.
\newblock A review of time-series anomaly detection techniques: A step to future perspectives.
\newblock In \emph{Advances in Information and Communication: Proceedings of the 2021 Future of Information and Communication Conference (FICC), Volume 1}, 865--877. Springer.

\bibitem[{Siuly, Li, and Zhang(2016)}]{siuly2016eeg}
Siuly, S.; Li, Y.; and Zhang, Y. 2016.
\newblock EEG signal analysis and classification.
\newblock \emph{IEEE Trans Neural Syst Rehabilit Eng}, 11: 141--144.

\bibitem[{Song et~al.(2020)Song, Lin, Guo, and Wan}]{song2020spatial}
Song, C.; Lin, Y.; Guo, S.; and Wan, H. 2020.
\newblock Spatial-temporal synchronous graph convolutional networks: A new framework for spatial-temporal network data forecasting.
\newblock In \emph{Proceedings of the AAAI conference on artificial intelligence}, volume~34, 914--921.

\bibitem[{Tan et~al.(2020)Tan, Saha, Jacoby, Henze, and Sarkar}]{tan2020granger}
Tan, S.~Y.; Saha, H.; Jacoby, M.; Henze, G.; and Sarkar, S. 2020.
\newblock Granger causality based hierarchical time series clustering for state estimation.
\newblock \emph{IFAC-PapersOnLine}, 53(2): 524--529.

\bibitem[{Thorndike(1953)}]{thorndike1953elbow}
Thorndike, R.~L. 1953.
\newblock Who belongs in the family?
\newblock \emph{Psychometrika}, 18(4): 267--276.

\bibitem[{Torres et~al.(2021)Torres, Hadjout, Sebaa, Mart{\'\i}nez-{\'A}lvarez, and Troncoso}]{torres2021deep}
Torres, J.~F.; Hadjout, D.; Sebaa, A.; Mart{\'\i}nez-{\'A}lvarez, F.; and Troncoso, A. 2021.
\newblock Deep learning for time series forecasting: a survey.
\newblock \emph{Big Data}, 9(1): 3--21.

\bibitem[{Ulanova, Begum, and Keogh(2015)}]{ulanova2015scalable}
Ulanova, L.; Begum, N.; and Keogh, E. 2015.
\newblock Scalable clustering of time series with u-shapelets.
\newblock In \emph{Proceedings of the 2015 SIAM international conference on data mining}, 900--908. SIAM.

\bibitem[{van~der Maaten and Hinton(2008)}]{JMLR:v9:vandermaaten08a}
van~der Maaten, L.; and Hinton, G. 2008.
\newblock Visualizing Data using t-SNE.
\newblock \emph{Journal of Machine Learning Research}, 9(86): 2579--2605.

\bibitem[{Veličković et~al.(2018)Veličković, Cucurull, Casanova, Romero, Liò, and Bengio}]{veličković2018graphattentionnetworks}
Veličković, P.; Cucurull, G.; Casanova, A.; Romero, A.; Liò, P.; and Bengio, Y. 2018.
\newblock Graph Attention Networks.
\newblock arXiv:1710.10903.

\bibitem[{Wu et~al.(2019)Wu, Souza, Zhang, Fifty, Yu, and Weinberger}]{wu2019simplifying}
Wu, F.; Souza, A.; Zhang, T.; Fifty, C.; Yu, T.; and Weinberger, K.~Q. 2019.
\newblock Simplifying Graph Convolutional Networks.
\newblock In \emph{Proceedings of the 36th International Conference on Machine Learning}, 6861--6871.

\bibitem[{Xi et~al.(2023)Xi, Jain, Zhang, and Lin}]{xi2023lb}
Xi, W.; Jain, A.; Zhang, L.; and Lin, J. 2023.
\newblock Lb-simtsc: An efficient similarity-aware graph neural network for semi-supervised time series classification.
\newblock \emph{arXiv preprint arXiv:2301.04838}.

\bibitem[{Xie, Girshick, and Farhadi(2016)}]{xie2016unsupervised}
Xie, J.; Girshick, R.; and Farhadi, A. 2016.
\newblock Unsupervised deep embedding for clustering analysis.
\newblock In \emph{International conference on machine learning}, 478--487. PMLR.

\bibitem[{Yang and Leskovec(2011)}]{yang2011patterns}
Yang, J.; and Leskovec, J. 2011.
\newblock Patterns of temporal variation in online media.
\newblock In \emph{Proceedings of the fourth ACM international conference on Web search and data mining}, 177--186.

\bibitem[{Yu, Yin, and Zhu(2017)}]{yu2017spatio}
Yu, B.; Yin, H.; and Zhu, Z. 2017.
\newblock Spatio-temporal graph convolutional networks: A deep learning framework for traffic forecasting.
\newblock \emph{arXiv preprint arXiv:1709.04875}.

\bibitem[{Zha et~al.(2022)Zha, Lai, Zhou, and Hu}]{zha2022towards}
Zha, D.; Lai, K.-H.; Zhou, K.; and Hu, X. 2022.
\newblock Towards similarity-aware time-series classification.
\newblock In \emph{Proceedings of the 2022 SIAM International Conference on Data Mining (SDM)}, 199--207. SIAM.

\bibitem[{Zhang et~al.(2016)Zhang, Wu, Yang, Tian, and Zhang}]{zhang2016unsupervised}
Zhang, Q.; Wu, J.; Yang, H.; Tian, Y.; and Zhang, C. 2016.
\newblock Unsupervised feature learning from time series.
\newblock In \emph{IJCAI}, 2322--2328. New York, USA.

\bibitem[{Zhang et~al.(2018)Zhang, Wu, Zhang, Long, and Zhang}]{zhang2018salient}
Zhang, Q.; Wu, J.; Zhang, P.; Long, G.; and Zhang, C. 2018.
\newblock Salient subsequence learning for time series clustering.
\newblock \emph{IEEE transactions on pattern analysis and machine intelligence}, 41(9): 2193--2207.

\bibitem[{Zhao et~al.(2020)Zhao, Wang, Duan, Huang, Cao, Tong, Xu, Bai, Tong, and Zhang}]{zhao2020multivariate}
Zhao, H.; Wang, Y.; Duan, J.; Huang, C.; Cao, D.; Tong, Y.; Xu, B.; Bai, J.; Tong, J.; and Zhang, Q. 2020.
\newblock Multivariate time-series anomaly detection via graph attention network.
\newblock In \emph{2020 IEEE international conference on data mining (ICDM)}, 841--850. IEEE.

\bibitem[{Zhao and Itti(2018)}]{zhao2018shapedtw}
Zhao, J.; and Itti, L. 2018.
\newblock shapeDTW: Shape dynamic time warping.
\newblock \emph{Pattern Recognition}, 74: 171--184.

\bibitem[{Zhao, Zhu, and Lu(2009)}]{zhao2009productivity}
Zhao, J.; Zhu, N.; and Lu, S. 2009.
\newblock Productivity model in hot and humid environment based on heat tolerance time analysis.
\newblock \emph{Building and environment}, 44(11): 2202--2207.

\end{thebibliography}

\end{document}